\documentclass[10pt,twocolumn,letterpaper]{article}

\usepackage{cvpr}
\usepackage{times}
\usepackage{epsfig}
\usepackage{graphicx}
\usepackage{amsmath}
\usepackage{amssymb}

\usepackage{siunitx}
\usepackage{cite}
\usepackage{ctable}
\usepackage{hhline}
\usepackage{booktabs}
\usepackage{subcaption}
\usepackage{csquotes }

\newcolumntype{L}[1]{>{\raggedright\let\newline\\\arraybackslash\hspace{0pt}}m{#1}}
\newcolumntype{C}[1]{>{\centering\let\newline\\\arraybackslash\hspace{0pt}}m{#1}}
\newcolumntype{R}[1]{>{\raggedleft\let\newline\\\arraybackslash\hspace{0pt}}m{#1}}

\usepackage[breaklinks=true,bookmarks=false]{hyperref}

\cvprfinalcopy 

\def\cvprPaperID{****} 

\ifcvprfinal\fi
\begin{document}

\title{PoseFix: Model-agnostic General Human Pose Refinement Network}

\author{Gyeongsik Moon\\
Department of ECE, ASRI\\
Seoul National University\\
{\tt\small mks0601@snu.ac.kr}
\and
Ju Yong Chang\\
Department of EI\\
Kwangwoon University\\
{\tt\small juyong.chang@gmail.com}
\and
Kyoung Mu Lee\\
Department of ECE, ASRI\\
Seoul National University\\
{\tt\small kyoungmu@snu.ac.kr}
}

\maketitle

\begin{abstract}
Multi-person pose estimation from a 2D image is an essential technique for human behavior understanding. In this paper, we propose a human pose refinement network that estimates a refined pose from a tuple of an input image and input pose. The pose refinement was performed mainly through an end-to-end trainable multi-stage architecture in previous methods. However, they are highly dependent on pose estimation models and require careful model design. By contrast, we propose a model-agnostic pose refinement method. According to a recent study, state-of-the-art 2D human pose estimation methods have similar error distributions. We use this error statistics as prior information to generate synthetic poses and use the synthesized poses to train our model. In the testing stage, pose estimation results of any other methods can be input to the proposed method. Moreover, the proposed model does not require code or knowledge about other methods, which allows it to be easily used in the post-processing step. We show that the proposed approach achieves better performance than the conventional multi-stage refinement models and consistently improves the performance of various state-of-the-art pose estimation methods on the commonly used benchmark. The code is available in this https URL\footnote{\url{https://github.com/mks0601/PoseFix_RELEASE}}.
\end{abstract}

\section{Introduction}

The goal of human pose estimation is to localize semantic keypoints of a human body. It is an essential technique for human behavior understanding and human-computer interaction. Recently, many methods~\cite{kocabas2018multiposenet,xiao2018simple,sun2017integral,he2017mask,chen2017cascaded,huang2017coarse,cao2016realtime,insafutdinov2016deepercut,papandreou2017towards,newell2017associative,pishchulin2016deepcut} utilize deep convolutional neural networks (CNNs) and achieved noticeable performance improvement. They are also updating performance limits in annual competitions for 2D human keypoint detection such as MS COCO keypoint detection challenge~\cite{lin2014microsoft}. 

In this paper, we propose a human pose refinement network that estimates a refined pose from a tuple of an input image and a pose. Conventionally, the pose refinement has been mainly performed by  multi-stage architectures ~\cite{newell2016stacked,wei2016convolutional,bulat2016human,chen2017cascaded}. In other words, the initial pose and image features generated in the first stage go through subsequent stages, and each stage outputs a refined pose. These multi-stage architectures are usually trained in an end-to-end manner. However, the conventional multi-stage architecture-based refinement approach is highly dependent on the pose estimation model and requires careful design for successful refinement. By contrast, in this work, we propose a model-agnostic pose refinement method that does not depend on the pose estimation model. 

\begin{figure}[t]
\begin{center}
   \includegraphics[width=1.0\linewidth]{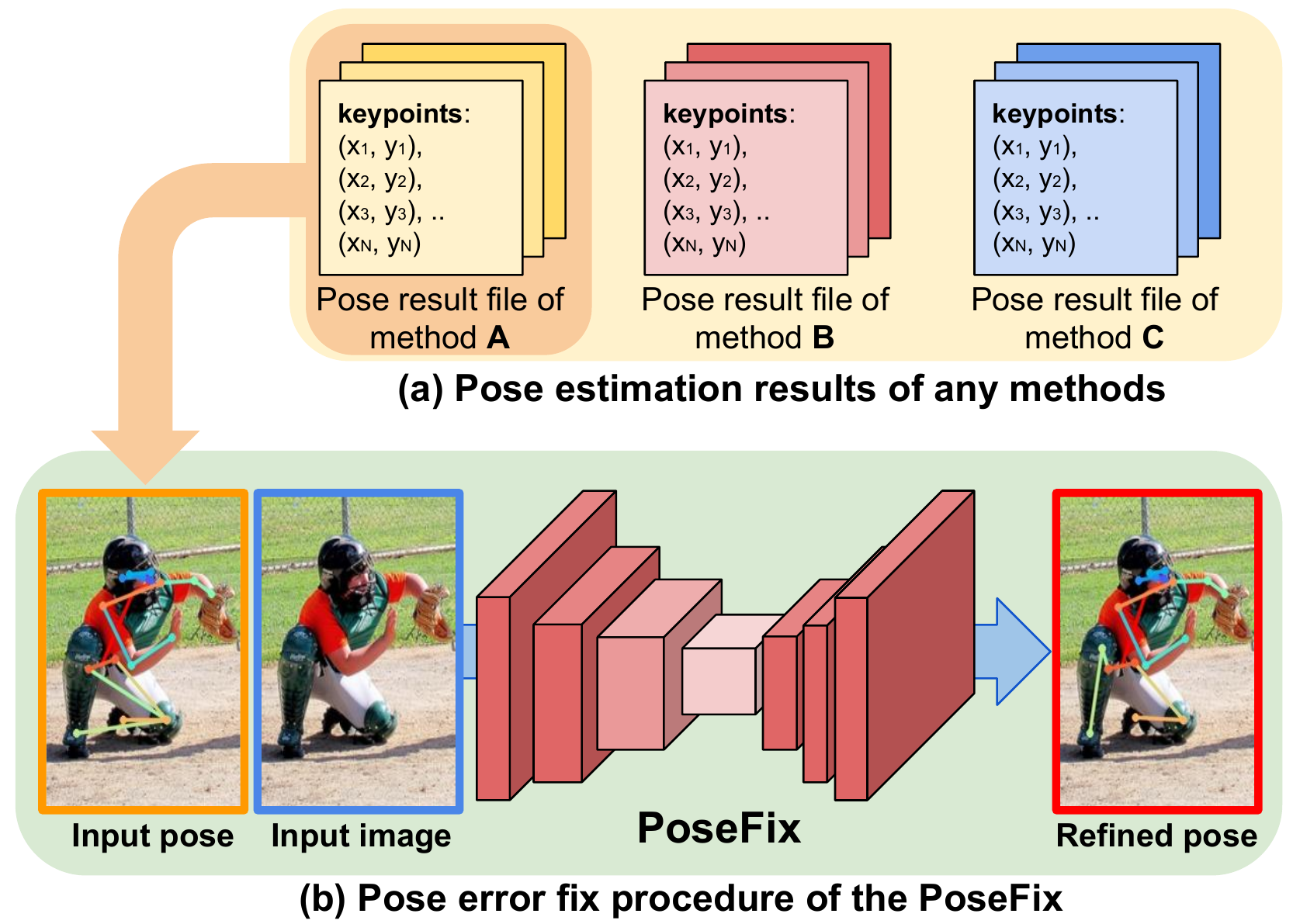}
\end{center}
\vspace*{-3mm}
   \caption{Testing pipeline of the PoseFix. It takes pose estimation results of any other method with an input image and outputs a refined pose. Note that the PoseFix does not require any code or knowledge about other methods.}
\vspace*{-3mm}
\label{fig:testing_pipeline}
\end{figure}

Recent research by Ronchi~\etal~\cite{ronchi2017benchmarking} gave us a clue on how to design a general model-agnostic pose refiner. They analyzed the results of the MS COCO 2016 keypoint detection challenge winners~\cite{cao2016realtime,papandreou2017towards} by using new pose estimation evaluation metrics, \textit{i.e.}, keypoint similarity (KS) and object keypoint similarity (OKS). They taxonomized pose estimation errors into several types such as \textit{jitter}, \textit{inversion}, \textit{swap}, and \textit{miss} and described how frequently these errors occur and how much they can negatively affect performance. Although the winners~\cite{cao2016realtime,papandreou2017towards} used very different approaches, their pose error distributions are very similar, which indicates that common issues exist for more accurate pose estimation. 

Our basic idea is to use this error statistics as prior information to generate synthetic poses and use the synthesized poses to train the proposed pose refinement model (PoseFix). To train our model, we generate each type of the errors (\textit{i.e.}, jitter, inversion, swap, and miss) based on the pose error distributions from Ronchi~\etal~\cite{ronchi2017benchmarking}, and construct diverse and realistic poses. The generated input pose is fed to the PoseFix with the input image, and the PoseFix learns to refine the pose. We design our PoseFix as a single-stage architecture with a coarse-to-fine estimation pipeline. It takes the input pose in a coarse form and estimates the refined pose in a finer form. The coarse input pose enables the proposed model to focus not only on an exact location of the input pose but also around it, allowing our model to fix the error of the input pose. Furthermore, the finer form of the output pose enables the proposed model to localize the location of the pose more exactly compared to existing methods. After training, our PoseFix can be applied to and refine the pose estimation results of any single- or multi-person pose estimation method. Figure~\ref{fig:testing_pipeline} shows such a pose refinement pipeline of the proposed PoseFix.


Our contributions can be summarized as follows.

\begin{itemize}
\item We show that model-agnostic general pose refinement is possible. The PoseFix is trained independently of the pose estimation model. Instead, it is based on error statistics obtained through empirical analysis.

\item  Our PoseFix can take the pose estimation result of any pose detection method as the input. As the PoseFix does not require any code or knowledge about other methods, our model has very high flexibility and accessibility.

\item We design the PoseFix as a coarse-to-fine estimation system. We empirically observed that this coarse-to-fine pipeline is crucial for successful pose refinement.

\item Our PoseFix achieves a better result than the conventional multi-stage architecture-based refinement methods. Also, the PoseFix consistently improves the performance of various state-of-the-art pose estimation methods on the commonly used benchmark.
\end{itemize}

\section{Related works}

\textbf{Single-person pose estimation.}
Toshev~\etal~\cite{toshev2014deeppose} directly estimated the Cartesian coordinates of body joints by using a multi-stage deep network and achieved state-of-the-art performance. Tompson~\etal~\cite{tompson2014joint} jointly trained a CNN and a graphical model. The CNN estimated 2D heatmaps for each joint, and they were used as the unary term for the graphical model. Liu~\etal~\cite{wei2016convolutional} used multi-stage CNN which progressively enlarges receptive fields and refines the pose estimation result. Newell~\etal~\cite{newell2016stacked} proposed a stacked hourglass network which repeats downsampling and upsampling to exploit multi-scale information effectively. Carreria~\etal~\cite{carreira2016human} proposed an iterative error feedback-based human pose estimation system. Chu~\etal~\cite{chu2017multi} enhanced the stacked hourglass network~\cite{newell2016stacked} by integrating it with a multi-context attention mechanism. Ke~\etal~\cite{ke2018multi} proposed a multi-scale structure-aware network which achieved leading position in the publicly available human pose estimation benchmark~\cite{andriluka20142d}.

\begin{figure*}
\begin{center}
\includegraphics[width=1.0\linewidth]{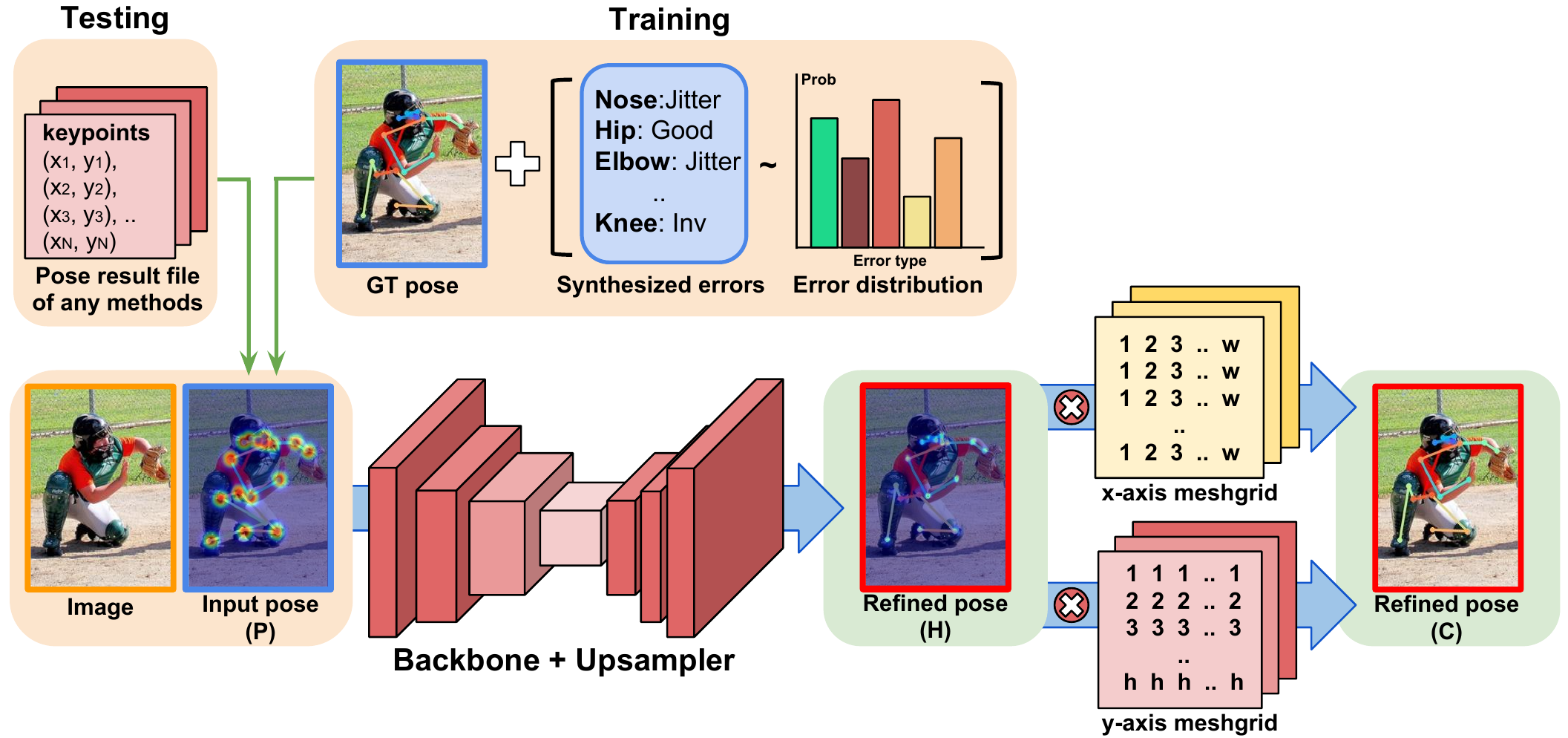}
\end{center}
\vspace*{-3mm}
   \caption{Overall pipeline of the PoseFix. In the training stage, the input pose is generated by synthesizing the pose errors based on the real pose error distributions on the groundtruth pose. In the testing stage, pose estimation results of any other methods become the input pose. The heatmaps are visualized by performing max pooling along the channel axis.}
\vspace*{-3mm}
\label{fig:PoseFix}
\end{figure*}

\textbf{Multi-person pose estimation.}
There are two main approaches in the multi-person pose estimation. The first one, top-down approach, relies on a human detector that predicts bounding boxes of humans. The detected human image is cropped and fed to the pose estimation network. The second one, bottom-up approach, localizes all human body keypoints in an input image and assembles them using proposed clustering algorithms in each work. 

~\cite{he2017mask,chen2017cascaded,huang2017coarse,papandreou2017towards,xiao2018simple,sun2017integral} are based on the top-down approach. He~\etal~\cite{he2017mask} proposed Mask R-CNN that can perform human detection and keypoint localization in a single model. Instead of cropping the detected humans in the input image, it crops human features from a feature map via the differentiable RoIAlign layer. Chen~\etal~\cite{chen2017cascaded} proposed a cascaded pyramid network (CPN) which consists of two networks. The first one, GlobalNet, is based on deep backbone network and upsampling layers with skip connections. The second one, RefineNet, is built to refine the estimation results from the GlobalNet by focusing on hard keypoints. Xiao~\etal~\cite{xiao2018simple} used a simple pose estimation network that consists of a deep backbone network and several upsampling layers. Although it is based on a simple network architecture, it achieved state-of-the-art performance on the commonly used benchmark~\cite{lin2014microsoft}.

~\cite{cao2016realtime,insafutdinov2016deepercut,newell2017associative,pishchulin2016deepcut,kocabas2018multiposenet} are based on the bottom-up approach. DeepCut~\cite{pishchulin2016deepcut} assigned the detected keypoints to each person in an image by formulating the assignment problem as an integer linear program. DeeperCut~\cite{pishchulin2016deepcut} improves the DeepCut~\cite{pishchulin2016deepcut} by introducing image-conditioned pair-wise terms. Cao~\etal~\cite{cao2016realtime} proposed part affinity fields (PAFs) that directly expose the association between human body keypoints. They assembled the localized keypoints of all persons in the input image by using the estimated PAFs. Newell~\etal~\cite{newell2017associative} introduced a pixel-wise tag value to assign localized keypoints to a certain human. Kocabas~\etal~\cite{kocabas2018multiposenet} proposed a pose residual network to assign detected keypoints to each person. Their model can jointly handle person detection, keypoint detection, and person segmentation.

\textbf{Human pose refinement.}
Many methods attempted to refine the estimated keypoint for more accurate performance. Newell~\etal~\cite{newell2016stacked}, Bulat and Tzimiropoulos~\cite{bulat2016human}, Liu~\etal~\cite{wei2016convolutional}, and Chen~\etal~\cite{chen2017cascaded} utilized an end-to-end trainable multi-stage architecture-based network. Each stage tries to refine the pose estimation results of the previous stage via end-to-end learning. Carreria~\etal~\cite{carreira2016human} iteratively estimated error feedback from a shared weight model. The output error feedback of the previous iteration is transformed into the input pose of the next iteration, which is repeated several times for progressive pose refinement. All of these methods combine pose estimation and refinement into a single model, and each refinement module is dependent on estimation. Therefore, the refinement modules have different structures, and they are not guaranteed to work successfully when they are combined with other estimation methods. On the other hand, our pose refinement method is independent of the estimation, and therefore the results can be consistently improved regardless of the prior pose estimation method.

Recently, Fieraru~\etal~\cite{fieraru2018learning} proposed a post-processing network to refine the pose estimation results of other methods, which is conceptually similar to ours. They synthesized pose for training and employed simple network architecture that estimates refined heatmaps and offset vectors for each joint. While their method follows ad-hoc rules to generate input pose, our method is based on actual error statistics obtained through empirical analysis. Also, our network with coarse-to-fine structure achieves a much stronger refinement performance than their simpler one.

\section{Overview of the proposed model}

The goal of the PoseFix is to refine the input 2D coordinates of the human body keypoints of all persons in an input image. To address this problem, our system is constructed based on the top-down pipeline which processes a tuple of a cropped human image and a given pose estimation result of that human instead of processing an entire image including multiple persons. In the training stage, the input pose is synthesized on the groundtruth pose realistically and diversely. In the testing stage, pose estimation results of any other methods can be the input pose to our system. The overall pipeline of the PoseFix is illustrated in Figure~\ref{fig:PoseFix}.

\section{Synthesizing poses for training}

To train the PoseFix, we generate synthesized poses using the groundtruth poses. As the PoseFix should cover different pose estimation results from various methods in the testing stage, synthesized poses need to be diverse and realistic. To satisfy these properties, we generate synthesized poses randomly based on the error distributions of real poses as described in ~\cite{ronchi2017benchmarking}. The distributions include the frequency of each pose error (\textit{i.e.}, jitter, inversion, swap, and miss) according to the joint type, number of visible keypoints, and overlap in the input image. There may also be joints that do not have any error, which should be synthesized very close to the groundtruth to simulate correct estimations. Ronchi~\etal~\cite{ronchi2017benchmarking} called this status \textit{good}. Considering most of the empirical distributions in~\cite{ronchi2017benchmarking}, we compute the probability that each joint will have one of the pose errors or be in the good status.

\begin{figure}[t]
\begin{center}
   \includegraphics[width=1.0\linewidth]{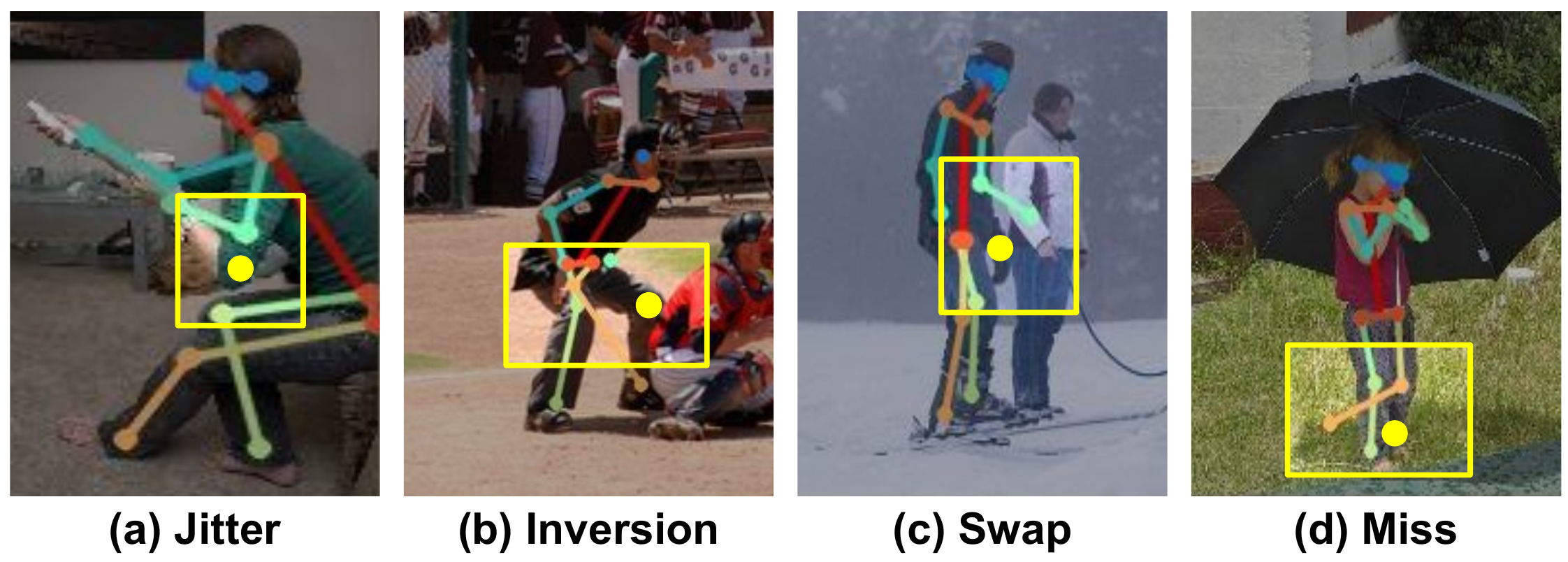}
\end{center}
\vspace*{-3mm}
   \caption{Visualization of synthesized pose errors for each type. The keypoint with pose error is highlighted by a yellow rectangle, and the groundtruth keypoints are drawn in a yellow circle.}
\vspace*{-3mm}   
\label{fig:synthesized_pose_error_type}
\end{figure}

The detailed error synthesis procedure on each groundtruth keypoint $\theta_j^p$ of joint $j$ which belongs to a person $p$ is described in below. For more clear description, we define $j'$ as a left/right inverted joint from the $j$, and $p'$ as a different person from the $p$ in the input image. Also, $d_j^k$ is defined as a $L2$ distance that makes KS with the groundtruth keypoint becomes $k$ for joint $j$. Note that $d_j^k$ depends on the type of joint $j$ because the error distribution of each joint has different scale ~\cite{ronchi2017benchmarking}. For example, eyes require more precise localization than hips to obtain the same KS $k$. Figure~\ref{fig:synthesized_pose_error_type} visualizes examples of synthesized pose errors of each type.

\textbf{Good.}
Good status is defined as a very small displacement from the groundtruth keypoint. An offset vector whose angle and length are uniformly sampled from [0, 2$\pi$) and [0, $d_j^{0.85}$), respectively, is added to the groundtruth $\theta_j^p$. The synthesized keypoint position should be closer to the original groundtruth $\theta_j^p$ than $\theta_{j'}^p$, $\theta_j^{p'}$, and $\theta_{j'}^{p'}$.

\textbf{Jitter.}
Jitter error is defined as a small displacement from the groundtruth keypoint. An offset vector whose angle and length are uniformly sampled from [0, 2$\pi$) and [$d_j^{0.85}$,$d_j^{0.5}$), respectively, is added to the groundtruth $\theta_j^p$. Similar to the \textit{good} status, the synthesized keypoint position should be closer to the original groundtruth $\theta_j^p$ than $\theta_{j'}^p$, $\theta_j^{p'}$, and $\theta_{j'}^{p'}$.

\textbf{Inversion.}
Inversion error occurs when a pose estimation model is confused between semantically similar parts that belong to the same instance. We restrict the inversion error to the left/right body part confusion following~\cite{ronchi2017benchmarking}. The \textit{jitter} error is added to $\theta_{j'}^p$. The synthesized keypoint position should be closer to the $\theta_{j'}^p$ than $\theta_j^p$, $\theta_j^{p'}$, and $\theta_{j'}^{p'}$.

\textbf{Swap.}
Swap error represents a confusion between the same or similar parts which belong to different persons. The \textit{jitter} is added to $\theta_j^{p'}$ or $\theta_{j'}^{p'}$. The closest keypoint from the synthesized keypoint should be $\theta_j^{p'}$ or $\theta_{j'}^{p'}$, not any of $\theta_j^p$ and $\theta_{j'}^p$.

\textbf{Miss.}
Miss error represents a large displacement from the groundtruth keypoint position. An offset vector whose angle and length are uniformly sampled from [0, 2$\pi$) and [$d_j^{0.5}$,$d_j^{0.1}$), respectively, is added to one of $\theta_j^p$, $\theta_j^{p'}$, $\theta_{j'}^p$, and $\theta_{j'}^{p'}$. The synthesized keypoint position should be at least $d_j^{0.5}$ away from all of $\theta_j^p$, $\theta_{j'}^p$, $\theta_j^{p'}$, and $\theta_{j'}^{p'}$.

\begin{figure}[t]
\begin{center}
   \includegraphics[width=1.0\linewidth]{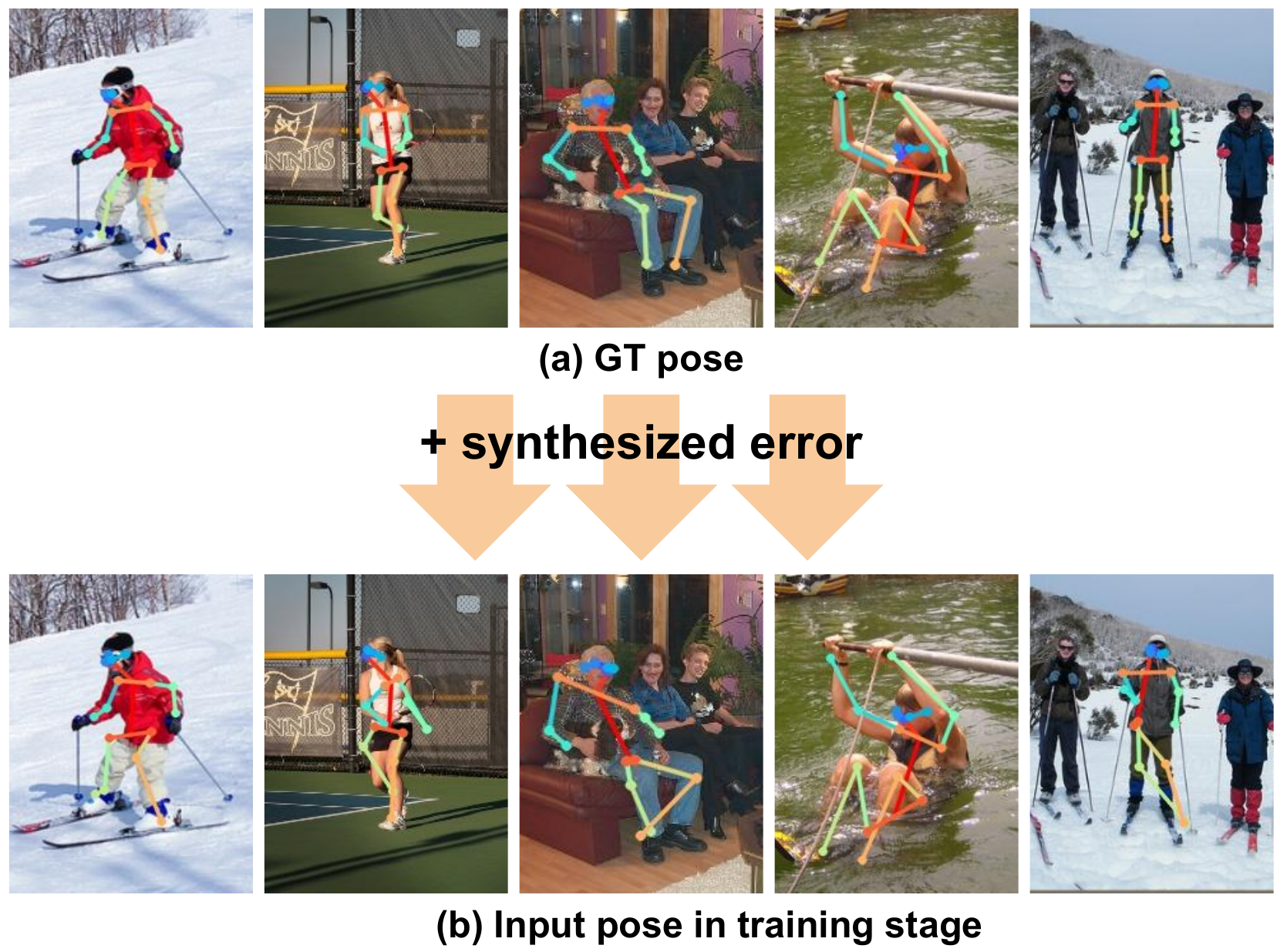}
\end{center}
\vspace*{-3mm}
   \caption{Visualization of the groundtruths and synthesized input poses. The synthesized poses are generated by adding errors to the groundtruth poses, which are used for training PoseFix.}
\vspace*{-3mm}   
\label{fig:input_pose}
\end{figure}

Some examples of synthesized input poses are shown in Figure~\ref{fig:input_pose}.

\section{Architecture and learning of PoseFix}

\subsection{Model design}

We design the PoseFix to directly estimate a refined pose from a tuple of an input image and an input pose as shown in Figure~\ref{fig:PoseFix}. The input image and the input pose provide contextual and structured information to the PoseFix, respectively, and the PoseFix learns to use these information to fix pose errors in the input pose. Although some errors exist in the input pose, it still provides useful structured information because, as indicated by Ronchi~\etal~\cite{ronchi2017benchmarking}, most keypoints in the input pose are in \textit{good} status or have \textit{jitter} error which represent a small displacement from the groundtruth pose. This rough structured information acts like attention which tells the PoseFix where to focus on at the human body. 

\begin{figure}[t]
\begin{center}
   \includegraphics[width=1.0\linewidth]{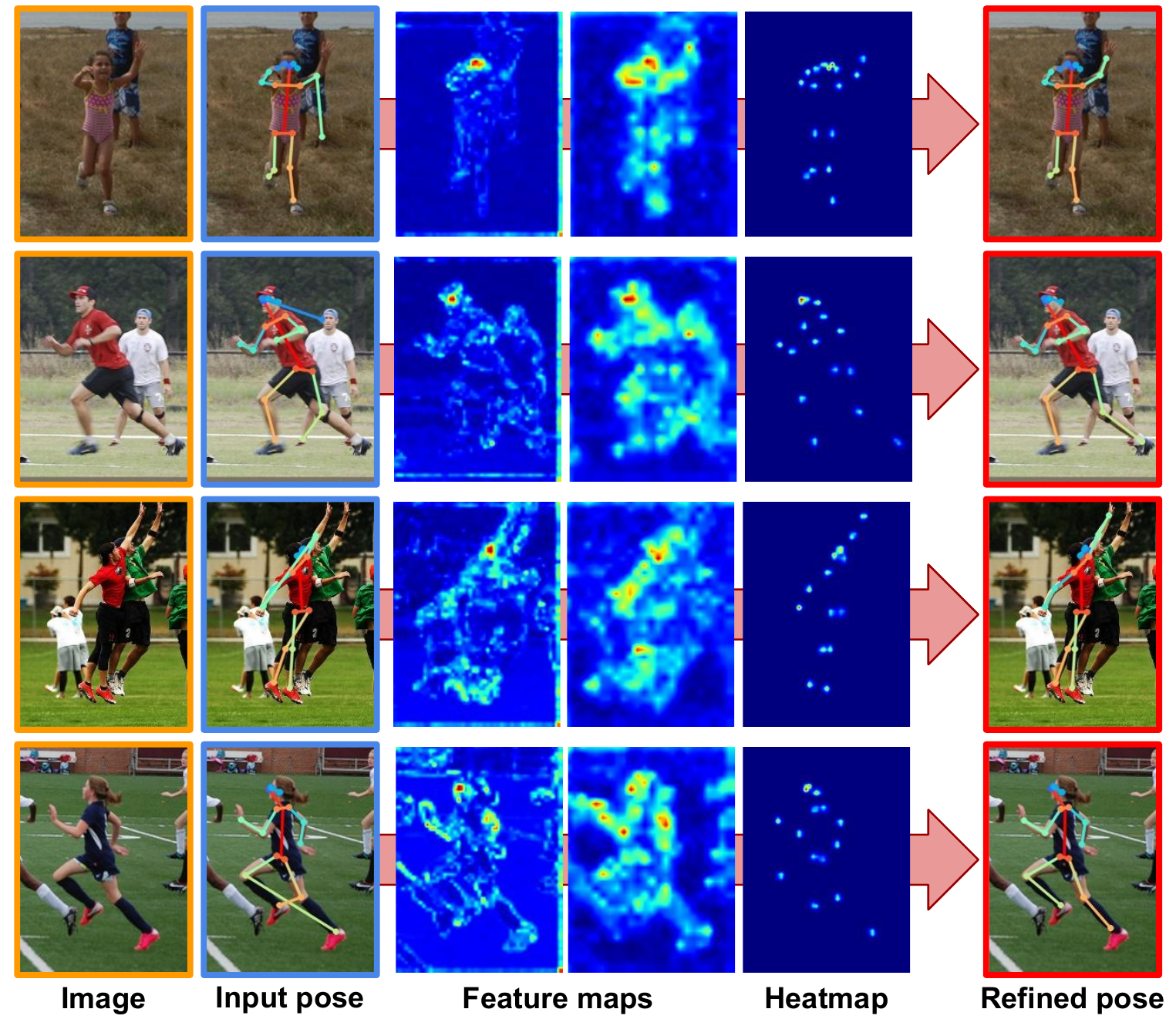}
\end{center}
\vspace*{-3mm}
   \caption{Visualization of feature maps and final heatmaps of the PoseFix. The feature maps and heatmaps are reduced into one channel by max pooling along the channel axis for visualization. The order of the feature maps and heatmaps in the figure is the same with that of the feedforward.}
\vspace*{-3mm}
\label{fig:intermediate_feature}
\end{figure}

We observed that by learning to fix the pose errors in the input pose, the PoseFix learns where to focus on at the human body as in Figure~\ref{fig:intermediate_feature}. As it shows, although some errors exist in the input pose, the PoseFix initially focuses well on the reliable keypoint locations of the input pose. And then, it successfully localizes correct keypoints without being influenced by the errors of the input pose.

\subsection{Coarse-to-fine estimation} \label{sec:c2f}

To make it more robust to errors, we design the proposed PoseFix to operate in a coarse-to-fine manner. We use the terms \enquote{coarse} and \enquote{fine} by the degree of uncertainty in representing the pose. For example, in representing the position of each joint constituting a pose, a Gaussian blob has a high uncertainty as much as the size of its standard deviation. On the other hand, a one-hot vector has relatively low uncertainty up to the size of a quantized grid. The coordinates of a keypoint has the least amount of uncertainty because it provides the exact information about the location itself. Therefore, in our work, the coarse-to-fine estimation implies that the coarse input pose ($P=\{P_n\}_{n=1}^{N}$) represented by the set of Gaussian blobs is fed to the network, producing the finer pose in the form of the one-hot vector ($H=\{H_n\}_{n=1}^{N}$), and then the finest pose in terms of the keypoint coordinates ($C=\{C_n\}_{n=1}^{N}$) is generated as the final output as illustrated in Figure~\ref{fig:PoseFix}. $N$ denotes the number of keypoints. In this subsection, we describe this coarse-to-fine estimation in more detail.


The input pose is constructed in a coarse form by a single-mode Gaussian heatmap representation as follows:
\begin{equation}
P_n (i,j) = \exp\left(-\frac{(i-i_{n})^2+(j-j_{n})^2}{2\sigma^2}\right),
\end{equation}
where $P_n$ and ($i_n$,$j_n$) are the input heatmap and 2D coordinates of $n$th keypoint, respectively, and $\sigma$ is the standard deviation of the Gaussian peak. The generated input pose is concatenated with the input image and fed into the PoseFix. This Gaussian heatmap representation is suitable for subsequent convolutional operations because it is pixel-wise aligned with the input image. Moreover, as the input pose can contain some errors, non-zero values around the center of the blob can be used to encourage the PoseFix to focus not only on the exact location of the input pose, but also around it.

From the input Gaussian heatmap in a coarse form, the proposed network generates the heatmap $H_n$ and the keypoint coordinates $C_n$ for the $n$th keypoint, sequentially. To make $H_n$ a finer form, we supervise it using a one-hot vector. Then, soft-argmax operation~\cite{sun2017integral} is applied to $H_n$ to generate $C_n$ in a differentiable manner. Soft-argmax is defined as the element-wise product between the input heatmap and the meshgrid followed by the summation, as shown in Figure~\ref{fig:PoseFix}. More precisely, the 2D coordinates are calculated from $H_n$ as follows:
\begin{equation}
C_n = {\left( \sum_{i=1}^{w}\sum_{j=1}^{h} iH_n (i,j), \sum_{i=1}^{w}\sum_{j=1}^{h} jH_n (i,j) \right)^{T}},
\end{equation}
where $w$ and $h$ are the width and height of $H_n$, respectively. Our network is trained by minimizing the cross-entropy-based integral loss~\cite{sun2017integral}, which is defined as follows:
\begin{equation}
L = L_H + L_C,
\end{equation}
where $L$ is the cross-entropy-based integral loss, and two losses $L_H$ and $L_C$ are described below.


The $L_H$ is a cross-entropy loss which is calculated after applying the softmax function to the output heatmap along the spatial axis. The definition of the $L_H$ is as follows: 
\begin{equation}
L _H= -\frac{1}{N}\sum_{n=1}^{N}\sum_{i,j}H_n^*(i,j)\log H_n (i,j),
\end{equation}
where $H_n^{*}$ and $H_n$ are the groundtruth and estimated heatmaps with softmax applied, respectively. The groundtruth heatmap $H_n^{*}$ is a one-hot vector if the groundtruth keypoint coordinates are integers. Otherwise, two grids for each $x$ and $y$ axis are selected by floor and ceil operations and are filled with probabilities by linear extrapolation. The $L_C$ is the sum of all $L1$ losses applied to the coordinates as follows:
\begin{equation}
L_C = \frac{1}{N} \sum_{n=1}^{N} \|C_{n}^*-C_{n}\|_1,
\end{equation}
where $C_{n}^{*}$ is the groundtruth coordinates vector for $n$th keypoint. The $L_H$ forces the PoseFix to select a single grid point in the estimated heatmap, and the $L_C$ enables the PoseFix to localize keypoints more precisely because it is calculated in the continuous space which is free from quantization errors. 


\subsection{Network architecture}
We used network architecture of Xiao~\etal~\cite{xiao2018simple} which consists of a deep backbone network (\textit{i.e.}, ResNet~\cite{he2016deep}) and several upsampling layers. The final upsampling layer becomes heatmaps ($H$) after applying the softmax function. The soft-argmax operation extracts coordinates ($C$) from the heatmaps ($H$), and it becomes the final estimation of the PoseFix.

\section{Implementation details} 

\textbf{Training.}
The proposed PoseFix is trained in an end-to-end manner. The weights of the backbone part are initialized with the publicly released ResNet model pre-trained on the ImageNet dataset~\cite{russakovsky2015imagenet}, and the weights of the remaining part are initialized from the zero-mean Gaussian distribution with $\sigma$ = 0.01 and as in He~\etal~\cite{he2015delving}. The weights are updated by Adam optimizer~\cite{kingma2014adam} with a mini-batch size of 128. The initial learning rate is set to 5$\times$$10^{-4}$ and reduced by a factor of 10 at 90 and 120th epoch. We perform data augmentation including scaling ($\pm$30\%), rotation ($\pm$\ang{40}), and flip. To crop humans from an input image, groundtruth human bounding boxes are extended to a fixed aspect ratio (\textit{i.e.}, height:width = 4:3) and then cropped without distorting the aspect ratio. The cropped bounding box is resized to a fixed size, which becomes the input image. We train the PoseFix 140 epochs with four NVIDIA 1080 Ti GPUs, which took two days.

\textbf{Testing.}
In the testing stage, the pose estimation result of other pose estimation methods becomes the input pose. To crop human bounding box from an image with multiple persons, we calculate bounding box coordinates from the keypoints coordinates of the input pose. Following ~\cite{chen2017cascaded,newell2016stacked}, we used testing time flip augmentation. 

Our model is implemented using TensorFlow~\cite{abadi2016tensorflow} deep learning framework.

\section{Experiment}

\subsection{Dataset and evaluation metric}
The proposed PoseFix is trained and tested on the MS COCO~\cite{lin2014microsoft} 2017 keypoint detection dataset, which consists of training, validation, and test-dev sets. The training set includes 57K images and 150K person instances. The validation set and the test-dev sets include 5K and 20K images, respectively. The OKS-based AP metric is used to evaluate the accuracy of the keypoint localization.

\begin{table}
\centering
\setlength\tabcolsep{1.0pt}
\def\arraystretch{1.1}
\begin{tabular}{L{2.7cm}C{1.0cm}C{1.0cm}C{1.0cm}C{1.0cm}C{1.0cm}}
\specialrule{.1em}{.05em}{.05em}
      Methods & $AP$ & $AP_{.50}$ & $AP_{.75}$ & $AP_M$ & $AP_L$ \\ \hline
E2E-refine    & 70.1 (+0.4) & 87.3 (-1.0)& 76.8 (-0.2) & 66.8 (+0.6) & 76.3 (+0.2) \\ 
\textbf{MA-refine (Ours)}      & \textbf{72.1 (+2.4)} & \textbf{88.5 (+0.2)} & \textbf{78.3 (+1.3)} & \textbf{68.6 (+2.4)} & \textbf{78.2 (+2.1)}  \\ \specialrule{.1em}{.05em}{.05em}
\end{tabular}
\vspace*{-3mm}
\caption{AP comparison between the conventional end-to-end trainable multi-stage refinement model (E2E-refine) and the proposed model-agnostic refinement model (MA-refine) on the validation set. The number in the parenthesis denotes the AP change from the input pose (\textit{i.e.}, CPN).}
\vspace*{-4mm}
\label{table:refine_compare}
\end{table}

\begin{figure}[t]
\begin{center}
   \includegraphics[width=1.0\linewidth]{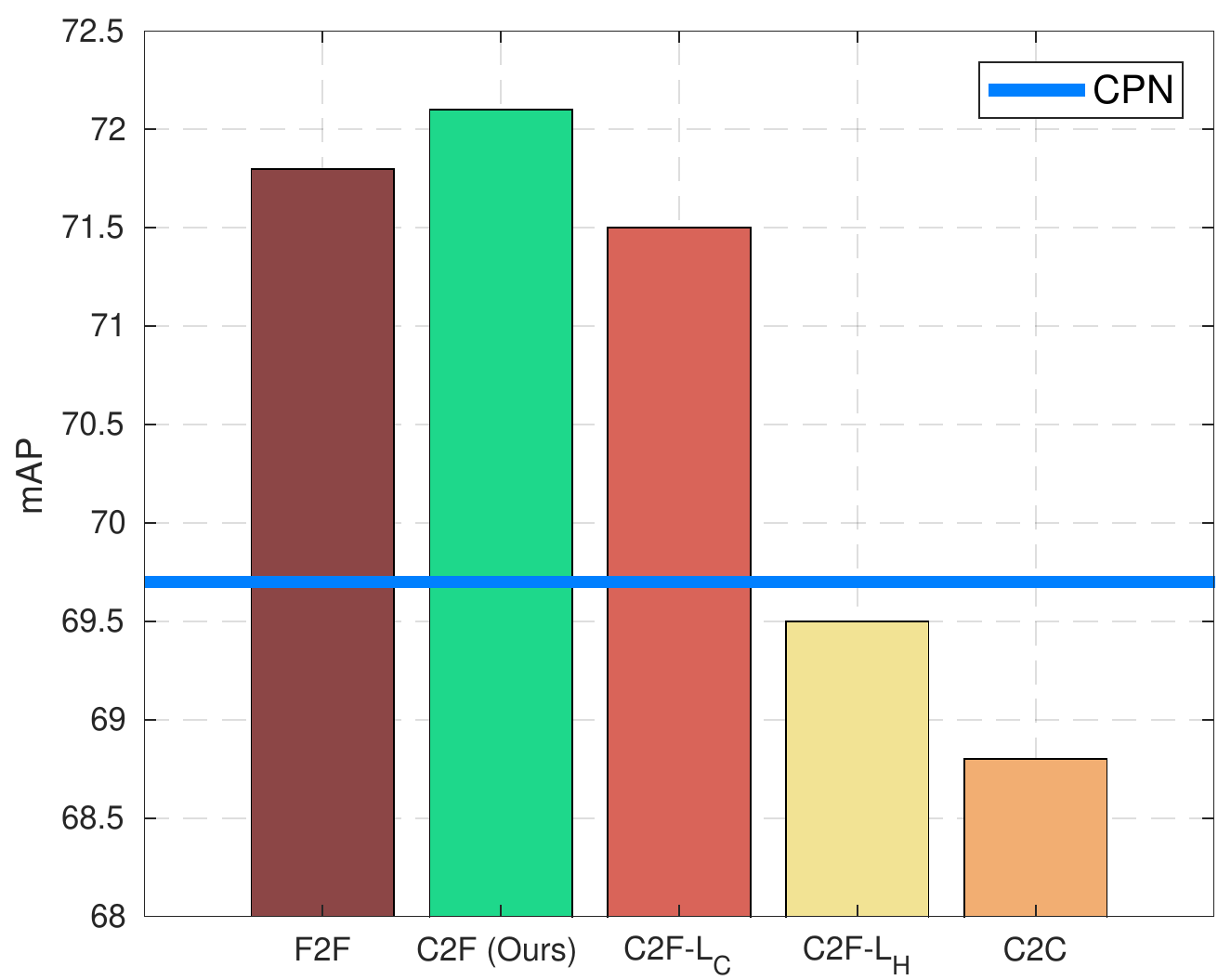}
\end{center}
\vspace*{-3mm}
   \caption{mAP comparison of various pipelines. The mAP is calculated on the validation set.}
\vspace*{-3mm}
\label{fig:input_output_type}
\end{figure}

\subsection{Ablation study}
To validate each component of the PoseFix, we tested the PoseFix on the validation set. The backbone of all the models are ResNet-50, and the size of the input image is set to 256$\times$192. We used the CPN~\cite{chen2017cascaded} which is a state-of-the-art human pose estimation method to generate the input poses.

\textbf{Model-agnostic pose refinement.}
We compared the accuracy of the conventional end-to-end trainable multi-stage architecture-based pose refinement model (E2E-refine) and the proposed model-agnostic refinement model (MA-refine) in Table~\ref{table:refine_compare}. To train the E2E-refine, we added a refinement module which has the same network architecture as the PoseFix at the end part of the pre-trained CPN. And then, we fine-tuned it by additionally giving the cross-entropy-based integral loss to the added module in an end-to-end manner. Both the input image and the output pose of the CPN are fed into the refinement module similarly to the PoseFix. We used a pre-trained CPN instead of training it from scratch because fine-tuning the pre-trained model yielded better performance.

\begin{figure}[t]
\begin{center}
   \includegraphics[width=1.0\linewidth]{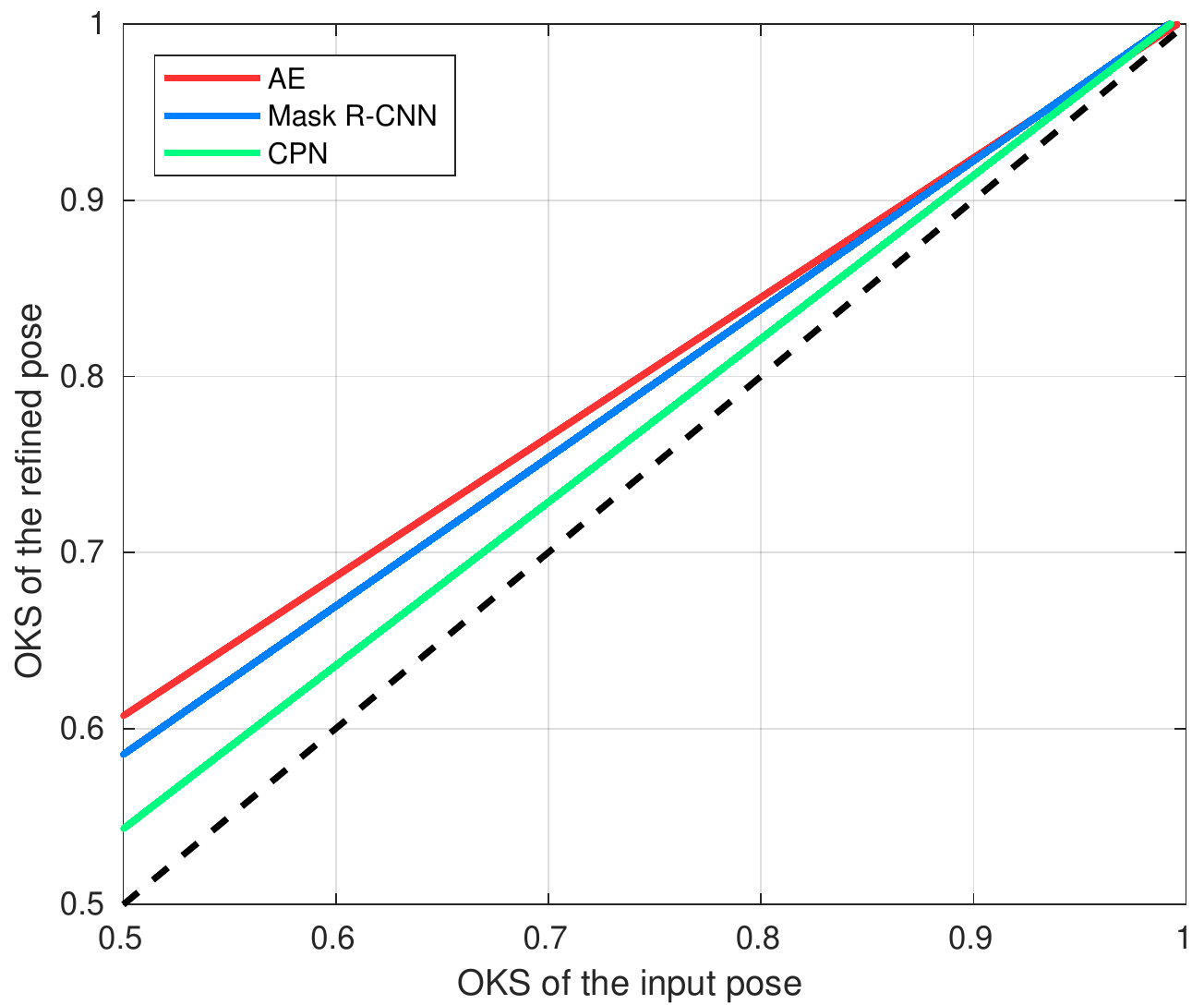}
\end{center}
\vspace*{-3mm}
   \caption{OKS change when the PoseFix is applied to state-of-the-art methods. The dotted line denotes identity function. OKS is calculated on the validation set.}
\vspace*{-3mm}
\label{fig:input_output_oks}
\end{figure}

\begin{figure}[t]
\begin{center}
   \includegraphics[width=1.0\linewidth]{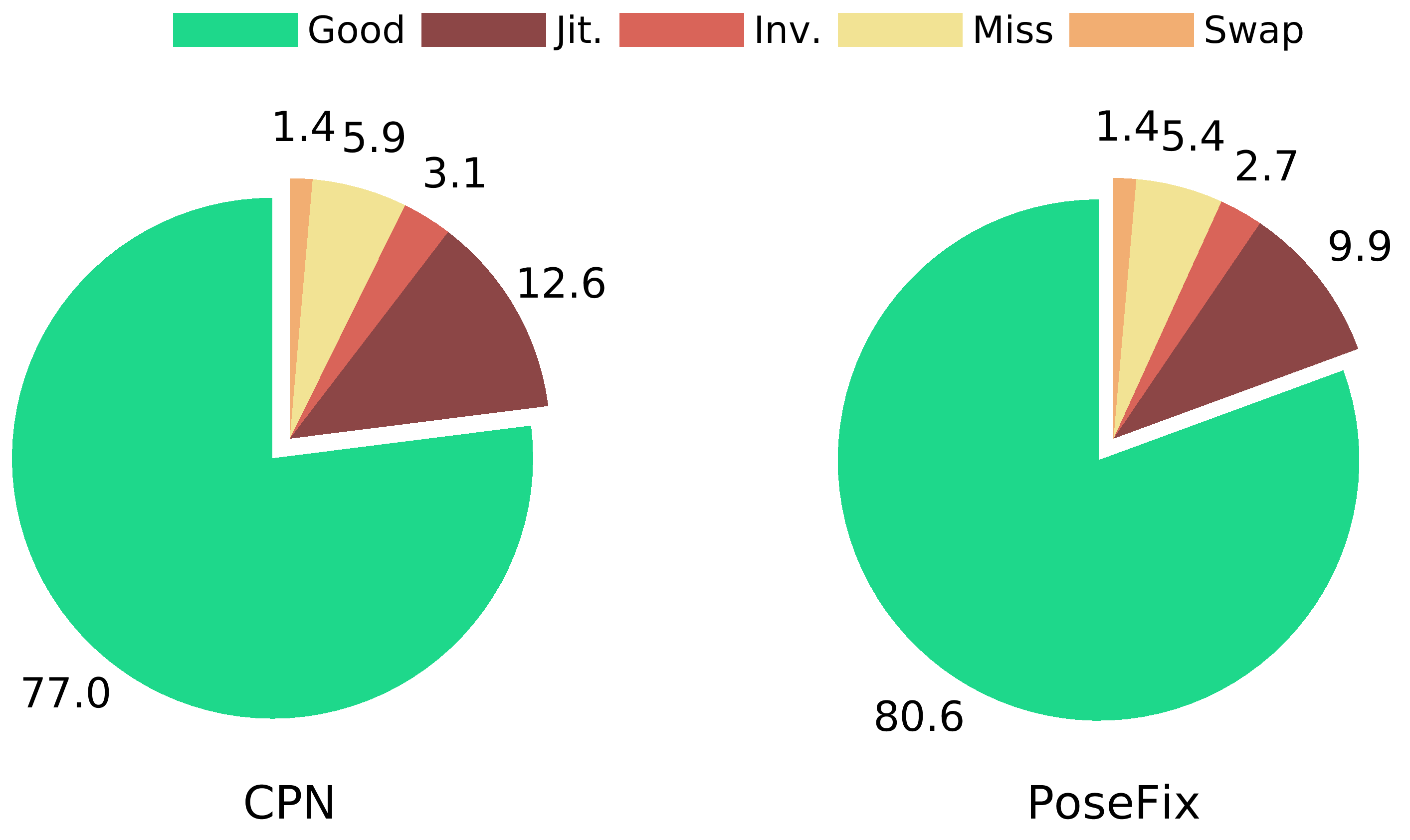}
\end{center}
\vspace*{-3mm}
   \caption{Frequency of each error type change when the PoseFix is applied to the CPN. The frequency is calculated on the validation set.}
\vspace*{-3mm}
\label{fig:input_output_error_freq}
\end{figure}

As Table~\ref{table:refine_compare} shows, the MA-refine trained in a model-agnostic manner improves the accuracy greatly more than the conventional refinement model does. We believe that this is because the added refinement module can be easily overfitted to the output pose of the CPN when training the E2E-refine. In contrast, various input poses that are realistically synthesized in the training stage of the PoseFix lead to the effect of data augmentation, which makes the PoseFix more robust to unseen input poses in the testing stage.

Instead of using the same network architecture of the PoseFix like in the E2E-refine, one can design their own refinement module. However, this approach requires careful network design because the amount of GPU memory available at one time is limited. By contrast, since the PoseFix is a decoupled model in both of the training and testing stages, it can serve as an add-on module, and thus provides more flexibility when building pose estimation models.

This analysis clearly demonstrates the benefits of using the model-agnostic pose refinement model compared with the conventional end-to-end trainable multi-stage architecture-based ones.

\begin{table*}
\centering
\setlength\tabcolsep{1.0pt}
\def\arraystretch{1.1}
\begin{tabular}{L{4.8cm}C{1.2cm}C{1.2cm}C{1.2cm}C{1.2cm}C{1.2cm}C{1.2cm}C{1.2cm}C{1.2cm}C{1.2cm}C{1.2cm}}
\specialrule{.1em}{.05em}{.05em}
      Methods & $AP$ & $AP_{.50}$ & $AP_{.75}$ & $AP_M$ & $AP_L$ & $AR$ & $AR_{.50}$ & $AR_{.75}$ & $AR_M$ & $AR_L$\\ \hline
AE~\cite{newell2017associative}    & 56.6 & 81.7 & 62.1 & 48.1 & 69.4 & 62.5 & 84.9 & 67.2 & 52.2 & 76.5 \\
+ \textbf{PoseFix (Ours)}    & \textbf{63.9} & \textbf{83.6} & \textbf{70.0} & \textbf{56.9} & \textbf{73.7} & \textbf{69.1} & \textbf{86.6} & \textbf{74.2} & \textbf{61.1} & \textbf{79.9} \\  \hline
PAFs~\cite{cao2016realtime}   & 61.7 & 84.9 & 67.4 & 57.1 & 68.1 & 66.5 & 87.2 & 71.7 & 60.5 & 74.6  \\ 
+ \textbf{PoseFix (Ours)}    & \textbf{66.7} & \textbf{85.7} & \textbf{72.9} & \textbf{62.9} & \textbf{72.3} & \textbf{71.3} & \textbf{88.0} & \textbf{76.7} & \textbf{66.3} & \textbf{78.1} \\  \hline
Mask R-CNN (ResNet-50)~\cite{he2017mask}  & 62.9 & 87.1 & 68.9 & 57.6 & 71.3 & 69.7 & 91.3 & 75.1 & 63.9 & 77.6 \\ 
+ \textbf{PoseFix (Ours)}     & \textbf{67.2} & \textbf{88.0} & \textbf{73.5} & \textbf{62.5} & \textbf{75.1} & \textbf{74.0} & \textbf{92.2} & \textbf{79.6} & \textbf{68.8} & \textbf{81.1} \\ 
Mask R-CNN (ResNet-101)  & 63.4 & 87.5 & 69.4 & 57.8 & 72.0 & 70.2 & 91.8 & 75.6 & 64.3 & 78.2 \\ 
+ \textbf{PoseFix (Ours)}    & \textbf{67.5} & \textbf{88.4} & \textbf{73.8} & \textbf{62.6} & \textbf{75.5} & \textbf{74.3} & \textbf{92.6} & \textbf{79.9} & \textbf{69.1} & \textbf{81.4} \\
Mask R-CNN (ResNeXt-101-64)  & 64.9 & 88.6 & 71.0 & 59.6 & 73.3 & 71.4 & 92.4 & 76.8 & 65.9 & 78.9 \\ 
+ \textbf{PoseFix (Ours)}     & \textbf{68.7} & \textbf{89.3} & \textbf{75.2} & \textbf{64.1} & \textbf{76.4} & \textbf{75.2} & \textbf{93.1} & \textbf{80.9} & \textbf{70.3} & \textbf{81.9} \\ 
Mask R-CNN (ResNeXt-101-32)  & 64.9 & 88.4 & 70.9 & 59.5 & 73.2 & 71.3 & 92.2 & 76.7 & 65.8 & 78.9 \\ 
+ \textbf{PoseFix (Ours)}    & \textbf{68.5} & \textbf{88.9} & \textbf{75.0} & \textbf{64.0} & \textbf{76.2} & \textbf{75.0} & \textbf{92.9} & \textbf{80.7} & \textbf{70.1} & \textbf{81.8} \\ \hline
IntegralPose & 66.3 & 87.6 & 72.9 & 62.7 & 72.7 & 73.2 & 91.8 & 79.1 & 68.3 & 79.8 \\ 
+ \textbf{PoseFix (Ours)} & \textbf{69.5} & \textbf{88.3} & \textbf{75.9} & \textbf{65.7} & \textbf{76.1} & \textbf{75.9} & \textbf{92.4} & \textbf{81.8} & \textbf{71.1} & \textbf{82.5} \\ \hline
CPN (ResNet-50)~\cite{chen2017cascaded} & 68.6 & 89.6 & 76.7 & 65.3 & 74.6 & 75.6 & 93.7 & 82.6 & 70.8 & 82.0 \\ 
+ \textbf{PoseFix (Ours)}    & \textbf{71.8} & \textbf{89.8} & \textbf{78.9} & \textbf{68.3} & \textbf{78.1} & \textbf{78.2} & \textbf{93.9} & \textbf{84.3} & \textbf{73.5} & \textbf{84.6} \\ 
CPN (ResNet-101) & 69.6 & 89.9 & 77.6 & 66.3 & 75.6 & 76.6 & 93.9 & 83.5 & 72.0 & 82.9 \\ 
+ \textbf{PoseFix (Ours)}   & \textbf{72.6} & \textbf{90.2} & \textbf{79.7} & \textbf{69.0} & \textbf{78.9} & \textbf{78.9} & \textbf{94.1} & \textbf{85.0} & \textbf{74.2} & \textbf{85.1} \\ \hline
Simple (ResNet-50)~\cite{xiao2018simple} & 69.4 & 90.1 & 77.4 & 66.2 & 75.5 & 75.1 & 93.9 & 82.4 & 70.8 & 81.0 \\ 
+ \textbf{PoseFix (Ours)} & \textbf{72.5} & \textbf{90.5} & \textbf{79.6} & \textbf{68.9} & \textbf{79.0} & \textbf{78.0} & \textbf{94.1} & \textbf{84.4} & \textbf{73.4} & \textbf{84.1} \\ 
Simple (ResNet-101) & 70.5 & 90.7 & 78.8 & 67.5 & 76.3 & 76.2 & 94.3 & 83.7 & 72.1 & 81.9 \\ 
+ \textbf{PoseFix (Ours)}  & \textbf{73.3} & \textbf{90.8} & \textbf{80.7} & \textbf{69.8} & \textbf{79.8} & \textbf{78.7} & \textbf{94.4} & \textbf{85.3} & \textbf{74.3} & \textbf{84.8} \\ 
Simple (ResNet-152) & 71.1 & 90.7 & 79.4 & 68.0 & 76.9 & 76.8 & \textbf{94.4} & 84.3 & 72.6 & 82.4 \\ 
+ \textbf{PoseFix (Ours)} & \textbf{73.6} & \textbf{90.8} & \textbf{81.0} & \textbf{70.3} & \textbf{79.8} & \textbf{79.0} & \textbf{94.4} & \textbf{85.7} & \textbf{74.8} & \textbf{84.9} \\  \specialrule{.1em}{.05em}{.05em}
\end{tabular}
\vspace*{-3mm}
\caption{Improvement of APs when the PoseFix is applied to the state-of-the-art methods. The APs are calculated on the test-dev set.}
\vspace*{-3mm}
\label{table:change_from_sota}
\end{table*}

\textbf{Coarse-to-fine estimation.}
To demonstrate the validity of the coarse-to-fine estimation, we compared the performance of fine-to-fine (\textit{i.e.}, $\mathrm{F2F}$), coarse-to-fine (\textit{i.e.}, $\mathrm{C2F}$, ours), and coarse-to-coarse (\textit{i.e.}, $\mathrm{C2C}$) estimation pipelines in Figure~\ref{fig:input_output_type}. As described in Section~\ref{sec:c2f}, the Gaussian heatmap and one-hot vector are used as coarse and fine forms of the input pose, respectively. To estimate the refined pose in a coarse form, the model learns to estimate the Gaussian heatmap by minimizing mean square error following~\cite{chen2017cascaded,newell2016stacked,wei2016convolutional}. For the fine-form estimation, cross-entropy-based integral loss is used as a loss function like ours. 

As Figure~\ref{fig:input_output_type} shows, the $\mathrm{C2F}$ (\textit{i.e.}, ours) exhibits a more accurate performance than $\mathrm{F2F}$, which indicates that coarse input pose representation is more beneficial than fine input pose representation. Also, $\mathrm{C2C}$ fails to improve the input pose whereas $\mathrm{F2F}$ and $\mathrm{C2F}$ successfully refine the input pose. These results indicate that the fine-form estimation is crucial for a successful refinement. 

To further analyze the benefit of the fine-form estimation, we additionally trained two models ($\mathrm{C2F}$-$L_H$ and $\mathrm{C2F}$-$L_C$). Instead of using both of the $L_C$ and $L_H$ like the $\mathrm{C2F}$ does, they are trained by minimizing only either $L_H$ or $L_C$. The $\mathrm{C2F}$-$L_H$ learns to estimate one-hot vector ($H$) by minimizing $L_H$, and $\mathrm{C2F}$-$L_C$ is supervised to estimate coordinate ($C$) by minimizing $L_C$. Among $\mathrm{C2C}$, $\mathrm{C2F}$-$L_H$, and $\mathrm{C2F}$-$L_C$, the target form of the $\mathrm{C2C}$ is the most coarse representation. On the other hand, that of $\mathrm{C2F}$-$L_C$ is the finest representation as described in Section~\ref{sec:c2f}. Figure~\ref{fig:input_output_type} shows that $\mathrm{C2C}$ yields the worst performance while $\mathrm{C2F}$-$L_C$ achieves the best among them. This finding shows that as the output representation of the PoseFix becomes a finer form, the performance improves. Thus, by integrating the two loss functions (\textit{i.e.}, $L_H$ and $L_C$) together, we can improve the performance much, as $\mathrm{C2F}$ shows.

This analysis clearly shows the benefit of the coarse-to-fine estimation pipeline.

\subsection{Performance improvement of the state-of-the-art methods by PoseFix}

We report the performance improvement when the PoseFix is applied to the recent state-of-the-art human pose estimation methods. PAFs~\cite{cao2016realtime}, AE~\cite{newell2017associative}, Mask R-CNN~\cite{he2017mask}, CPN~\cite{chen2017cascaded}, and Simple~\cite{xiao2018simple} are used to generate the input pose. To obtain the pose estimation results of the previous methods, we used their released codes and pre-trained models. We tested them by ourselves without ensembling and testing time augmentation. We also trained a pose estimation model (IntegralPose) with the same network architecture and loss function with the PoseFix to show that the PoseFix can improve a model trained from the same architecture. To analyze how the PoseFix changes the OKS and frequency of each error type, we tested the PoseFix on the validation set. We also report how much the PoseFix improves AP on the test-dev set. The ResNet-152 is used as the backbone of the PoseFix, and the size of the input image is set to 384$\times$288.

\textbf{OKS change.}
The graph in Figure~\ref{fig:input_output_oks} shows the change of the OKS of the same instance when the PoseFix is applied to the baseline state-of-the-art methods.

\textbf{Error frequency change.}
Figure~\ref{fig:input_output_error_freq} shows how the frequency of each status or error type changes when the PoseFix is applied to the CPN.

\textbf{AP improvement.}
Table~\ref{table:change_from_sota} shows the improvements in AP when the PoseFix is applied to the recent state-of-the-art human pose estimation methods. We also included the results of using different backbone networks~\cite{he2016deep,xie2017aggregated} for the Mask R-CNN, CPN, and Simple.

As Figures~\ref{fig:input_output_oks},~\ref{fig:input_output_error_freq} and Table~\ref{table:change_from_sota} show, the PoseFix consistently improves the performance of the state-of-the-art methods. The PoseFix corrects not only the small displacement error (\textit{i.e.}, jitter), but also the large displacement errors (\textit{i.e.}, inversion, miss, and swap) as in Figure~\ref{fig:input_output_error_freq}. Taking into account the fact that the state-of-the-art methods used in the experiments vary in structure and learning strategies, we believe that our model has generalizability that can be applied to other pose estimation methods. It is also noticeable that the PoseFix does not require any code or knowledge of the pose estimation methods, which makes our model very easy and convenient to use in practice.

\section{Conclusion}

We proposed a novel and powerful network, PoseFix, for human pose refinement. Unlike conventional end-to-end multi-stage architecture models, the proposed PoseFix is a model-agnostic pose refinement network. To train the PoseFix, we generate the input pose by synthesizing pose errors according to empirical pose error distributions on the groundtruth pose. The PoseFix takes an input pose in a coarse form and estimates the refined pose in a finer form. Since PoseFix is model-agnostic, it does not require any code or knowledge about the target models. So, it can be used as a post-processing add-on module conveniently. We showed that the PoseFix achieves better performance than the conventional multi-stage architecture-based pose refinement module. Furthermore, the PoseFix consistently improves the accuracy of other methods on the commonly used pose estimation benchmark.

\clearpage

\begin{center}
\textbf{\large Supplementary Material of \enquote{PoseFix: Model-agnostic General Human Pose Refinement Network}}
\end{center}

\setcounter{section}{0}

In this supplementary material, we present more experimental results that could not be included in the main manuscript due to the lack of space.

\section{Comparison with conventional end-to-end trainable multi-stage refinement}
In Table~\ref{table:refine_compare} of the main manuscript, we compared the accuracy of the conventional end-to-end trainable multi-stage refinement model (E2E-refine) and the proposed model-agnostic refinement model (MA-refine). We tried to show the effectiveness of the proposed model-agnostic refinement model by making the number of parameters of the E2E-refine and MA-refine same. 

However, as the conventional refinement requires careful model design, simply adding a refinement module which has the same network architecture with the PoseFix can result in sub-optimal performance. Therefore, we compare the accuracy of the refinement module of the state-of-the-art refinement-based method (\textit{i.e.}, CPN~\cite{chen2017cascaded}) and the PoseFix. The CPN consists of two parts. The first one, GlobalNet, is the baseline of the CPN. The second one, RefineNet, refines the pose estimation results of the GlobalNet. We use the GlobalNet as the pose estimation model and compare the accuracy improvement of the RefineNet and PoseFix. We trained and tested the CPN with GlobalNet only and both of the GlobalNet and RefineNet, using their released code.

Table~\ref{table:refine_compare_2} shows our PoseFix improves AP more than state-of-the-art refinement module (\textit{i.e.}, RefineNet) by a large margin. This comparison demonstrates the benefit of the model-agnostic refinement over conventional end-to-end trainable multi-stage refinement more clearly.

\section{Performance improvement of the state-of-the-art methods by PoseFix}
\begin{figure}[t]
\begin{center}
   \includegraphics[width=1.0\linewidth]{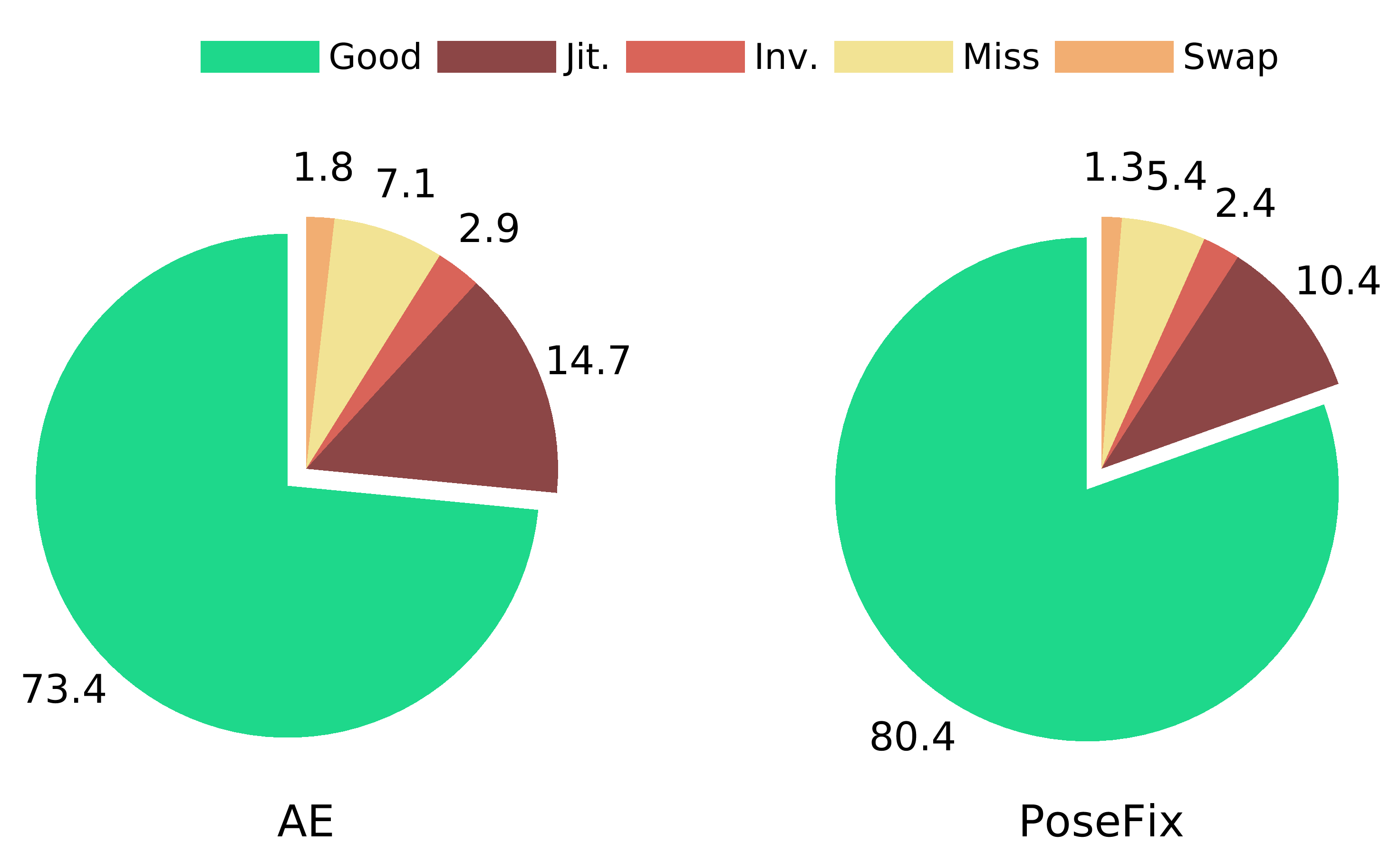}
\end{center}
\vspace*{-3mm}
   \caption{Frequency of each error type change when the PoseFix is applied to the AE. The frequency is calculated on the MS COCO~\cite{lin2014microsoft} validation set.}
\vspace*{-3mm}
\label{fig:AE_change}
\end{figure}

\begin{figure}[t]
\begin{center}
   \includegraphics[width=1.0\linewidth]{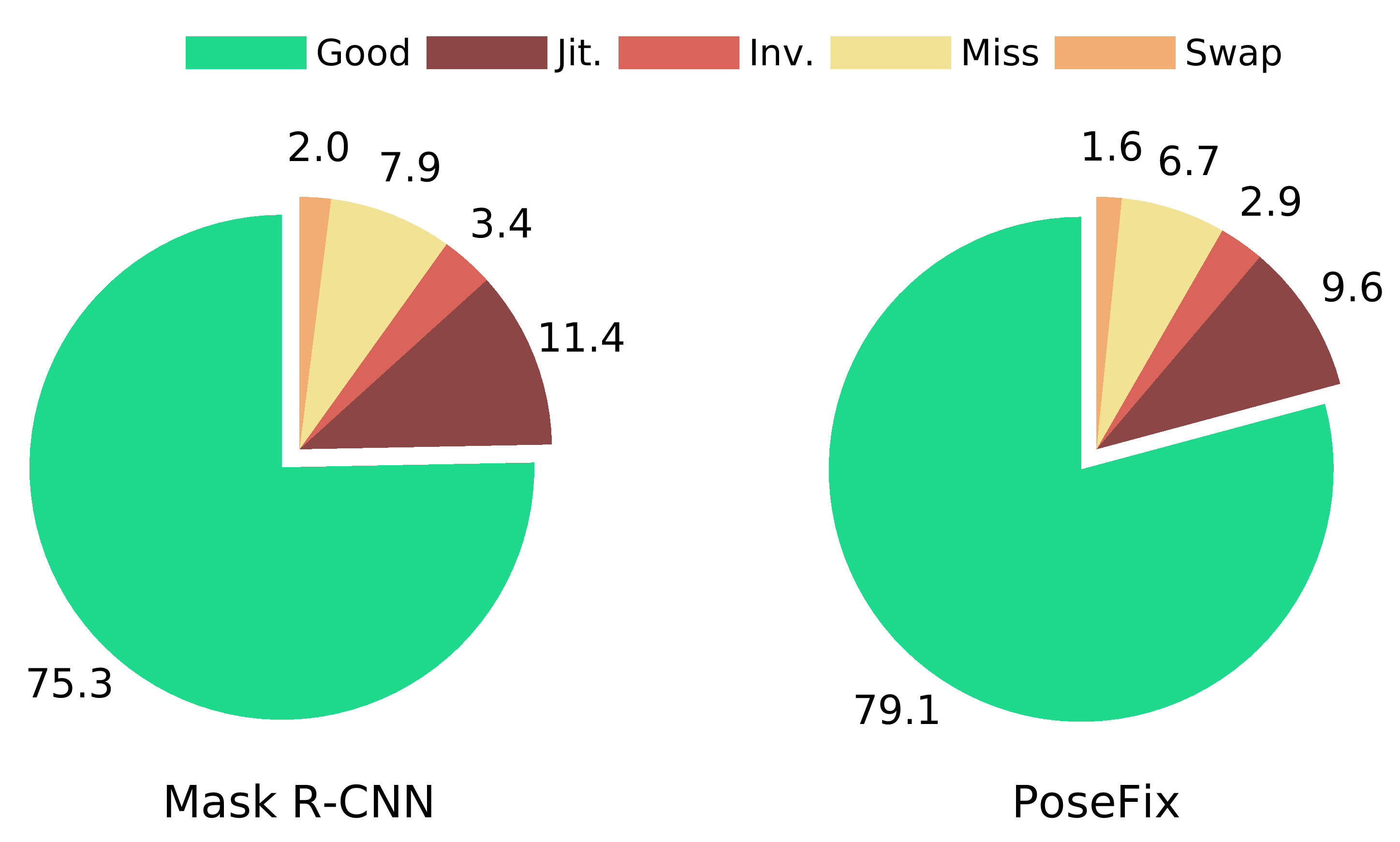}
\end{center}
\vspace*{-3mm}
   \caption{Frequency of each error type change when the PoseFix is applied to the Mask R-CNN. The frequency is calculated on the MS COCO~\cite{lin2014microsoft} validation set.}
\vspace*{-3mm}
\label{fig:MaskRCNN_change}
\end{figure}

In Figure~\ref{fig:input_output_error_freq} of the main manuscript, we showed how the frequency of each error type changes when the PoseFix is applied to the state-of-the-art method (\textit{i.e.}, CPN~\cite{chen2017cascaded}). We additionally show the changes of the AE~\cite{newell2017associative} and Mask R-CNN~\cite{he2017mask} in Figure~\ref{fig:AE_change} and ~\ref{fig:MaskRCNN_change}, respectively. As the Figures show, our PoseFix improves the performance by fixing all types of pose errors.

\begin{table}
\centering
\setlength\tabcolsep{1.0pt}
\def\arraystretch{1.1}
\begin{tabular}{L{2.7cm}C{1.0cm}C{1.0cm}C{1.0cm}C{1.0cm}C{1.0cm}}
\specialrule{.1em}{.05em}{.05em}
      Methods & $AP$ & $AP_{.50}$ & $AP_{.75}$ & $AP_M$ & $AP_L$ \\ \hline
RefineNet~\cite{chen2017cascaded}    & 69.1 (+1.8) & 87.9 (+0.4)& 76.6 (+2.2) & 65.7 (+1.6) & 75.5 (+2.2) \\ 
\textbf{PoseFix (Ours)}      & \textbf{71.5 (+4.2)} & \textbf{88.0 (+0.5)} & \textbf{77.6 (+3.2)} & \textbf{68.0 (+3.9)} & \textbf{78.1 (+4.8)}  \\ \specialrule{.1em}{.05em}{.05em}
\end{tabular}
\vspace*{-2mm}
\caption{AP comparison between state-of-the-art conventional end-to-end trainable multi-stage refinement model (RefineNet~\cite{chen2017cascaded}) and the proposed model-agnostic refinement model (PoseFix) on the MS COCO~\cite{lin2014microsoft} validation set. The number in the parenthesis denotes the AP change from the input pose (\textit{i.e.}, GlobalNet of the CPN~\cite{chen2017cascaded}).}
\vspace*{-4mm}
\label{table:refine_compare_2}
\end{table}

\begin{table*}
\centering
\setlength\tabcolsep{1.0pt}
\def\arraystretch{1.1}
\begin{tabular}{L{2.7cm}C{1.2cm}C{1.2cm}C{1.2cm}C{1.2cm}C{1.2cm}C{1.2cm}C{1.2cm}C{1.2cm}C{1.2cm}C{1.2cm}}
\specialrule{.1em}{.05em}{.05em}
      Methods & $AP$ & $AP_{.50}$ & $AP_{.75}$ & $AP_M$ & $AP_L$ & $AR$ & $AR_{.50}$ & $AR_{.75}$ & $AR_M$ & $AR_L$\\ \hline
RMPE~\cite{fang2017rmpe}  & 61.0 & 82.9 & 68.8 & 57.9 & 66.5 & - & - & - & - & - \\ 
PAFs~\cite{cao2016realtime}   & 61.8 & 84.9 & 67.5 & 57.1 & 68.2 & 66.5 & 87.2 & 71.8 & 60.6 & 74.6  \\ 
Mask R-CNN~\cite{he2017mask}  & 63.1 & 87.3 & 68.7 & 57.8 & 71.4 & - & - & - & - & - \\ 
AE~\cite{newell2017associative}    & 65.5 & 86.8 & 72.3 & 60.6 & 72.6 & 70.2 & 89.5 & 76.0 & 64.6 & 78.1 \\
Integral~\cite{sun2017integral} & 67.8 & 88.2 & 74.8 & 63.9 & 74.0 & - & - & - & - & - \\ 
G-RMI~\cite{papandreou2017towards}   & 64.9 & 85.5 & 71.3 & 62.3 & 70.0 & 69.7 & 88.7 & 75.5 & 64.4 & 77.1  \\ 
G-RMI*~\cite{papandreou2017towards}   & 68.5 & 87.1 & 75.5 & 65.8 & 73.3 & 73.3 & 90.1 & 79.5 & 68.1 & 80.4  \\ 
MultiPoseNet~\cite{kocabas2018multiposenet} & 69.6 & 86.3 & 76.6 & 65.0 & 76.3 & 73.5 & 88.1 & 79.5 & 68.6 & 80.3 \\ 
CFN~\cite{huang2017coarse}  & 72.6 & 86.1 & 69.7 & \textbf{78.3} & 64.1 & - & - & - & - & - \\ 
CPN~\cite{chen2017cascaded} & 72.1 & 91.4 & 80.0 & 68.7 & 77.2 & 78.5 & \textbf{95.1} & 85.3 & 74.2 & 84.3 \\ 
CPN++~\cite{chen2017cascaded} & 73.0 & 91.7 & 80.9 & 69.5 & 78.1 & 79.0 & \textbf{95.1} & 85.9 & 74.8 & 84.7 \\ 
Simple~\cite{xiao2018simple} & 73.7 & \textbf{91.9} & 81.1 & 70.3 & 80.0 & 79.0 & - & - & - & - \\ \hline
Simple~\cite{xiao2018simple} & 73.3 & 91.2 & 80.9 & 69.8 & 79.7 & 78.7 & 94.8 & 85.4 & 74.2 & 84.8 \\
\textbf{+ PoseFix (Ours)} & \textbf{74.9} & 91.2 & \textbf{81.9} & 71.1 & \textbf{81.2} & \textbf{79.9} & 94.8 & \textbf{86.3} & \textbf{75.5} & \textbf{86.0} \\ \hline
\end{tabular}
\vspace*{-2mm}
\caption{Comparison of APs with the state-of-the-art methods on the test-dev set. \enquote{*} means that the method involves extra data for training. \enquote{++} indicates results using ensemble.}
\vspace*{-4mm}
\label{table:comparison_with_sota}
\end{table*}

\begin{figure}[t]
\begin{center}
   \includegraphics[width=1.0\linewidth]{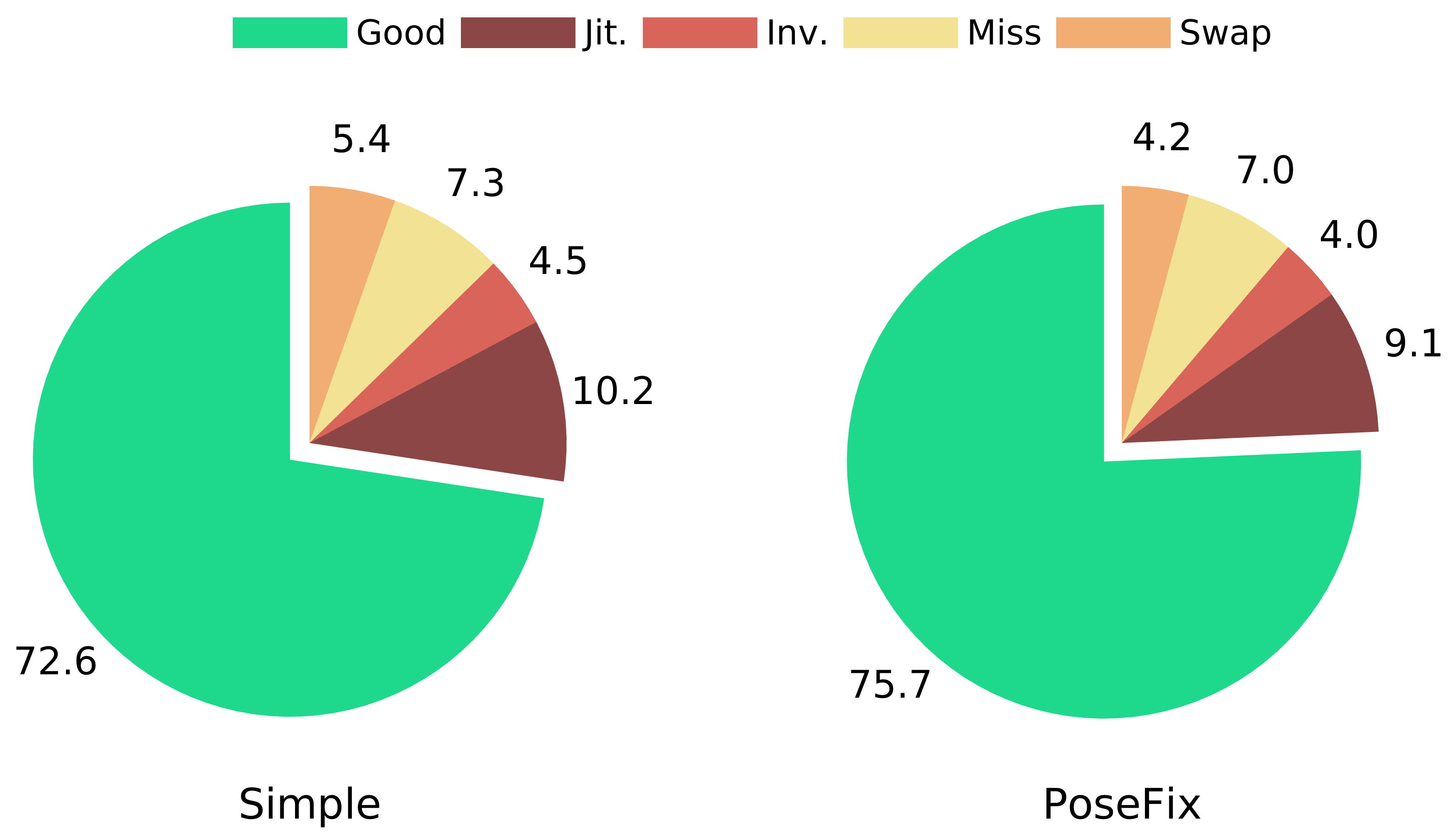}
\end{center}
\vspace*{-3mm}
   \caption{Frequency of each error type change when the PoseFix is applied to the Simple. The frequency is calculated on the PoseTrack 2018~\cite{andriluka2018posetrack} validation set.}
\vspace*{-5mm}
\label{fig:Simple_PoseTrack_change}
\end{figure}

We demonstrate more generalizability by showing performance improvement on another 2D multi-person pose estimation dataset (\textit{i.e.},  PoseTrack 2018~\cite{andriluka2018posetrack}). The PoseTrack 2018 dataset includes 66K frames, and they are split into training, validation and testing set. The state-of-the-art human pose estimation method, Simple~\cite{xiao2018simple}, is re-implemented by ours \footnote{\url{https://github.com/mks0601/TF-SimpleHumanPose}} and its testing result is used as the input pose of the PoseFix. Both of the Simple~\cite{xiao2018simple} and PoseFix is pre-trained on the COCO dataset and trained again on the PoseTrack 2018 training set without hyperparameter changes following~\cite{xiao2018simple}. Figure~\ref{fig:Simple_PoseTrack_change} and Table~\ref{table:change_from_sota_posetrack} show performance improvement on the PoseTrack 2018 validation set. As they show, the PoseFix significantly improves the performance of the input pose. They show the proposed PoseFix can improve the input pose on variable datasets.

\section{Comparison with state-of-the-art methods}
We compare the performance of the PoseFix with state-of-the-art methods, which include PAFs~\cite{cao2016realtime}, G-RMI~\cite{papandreou2017towards}, AE~\cite{newell2017associative}, RMPE~\cite{fang2017rmpe}, Mask R-CNN~\cite{he2017mask}, CFN~\cite{huang2017coarse}, CPN~\cite{chen2017cascaded}, Integral~\cite{sun2017integral}, MultiPoseNet~\cite{kocabas2018multiposenet}, and Simple~\cite{xiao2018simple} on the MS COCO~\cite{lin2014microsoft} test-dev set. All the performance are from their papers. We used Simple~\cite{xiao2018simple} as the input pose of the PoseFix. As they did not release the human detection model and result, we used our human detection model which achieves 57.2 AP for the human category on the test-dev set. The Simple~\cite{xiao2018simple} with our human detection model outputs slightly worse performance (73.3 AP) than the original Simple~\cite{xiao2018simple} (73.7 AP).

As shown in Table~\ref{table:comparison_with_sota}, our PoseFix outperforms all existing methods. It is noticeable that our method can achieve better performance when a new state-of-the-art method is proposed by using it as the input pose of our method. We also compare the performance of the PoseFix with PoseRefiner~\cite{fieraru2018learning} which has a similar approach to ours in Table~\ref{table:comparison_poserefiner}. Table shows our PoseFix improve input pose significantly more than PoseRefiner~\cite{fieraru2018learning}.

\begin{table}
\centering
\setlength\tabcolsep{1.0pt}
\def\arraystretch{1.1}
\begin{tabular}{C{1.7cm}C{0.8cm}C{0.8cm}C{0.7cm}C{0.7cm}C{0.7cm}C{0.8cm}C{0.8cm}C{0.8cm}}
\specialrule{.1em}{.05em}{.05em}
      Methods & Head & Shou & Elb & Wri & Hip & Knee & Ankl & Total \\ \hline
Simple~\cite{xiao2018simple}    & 74.4  & 76.9 & 72.2 & 65.2 & 69.2 & 70.0 & 62.9 & 70.4 \\ 
\textbf{+ PoseFix (Ours)}      & \textbf{79.0} & \textbf{81.6} & \textbf{76.4} & \textbf{69.7} & \textbf{75.2} & \textbf{74.3} & \textbf{67.0} & \textbf{75.0}\\ \specialrule{.1em}{.05em}{.05em}
\end{tabular}
\vspace*{-2mm}
\caption{Improvement of APs when the PoseFix is applied to the state-of-the-art method. The APs are calculated on the PoseTrack 2018 validation set.}
\vspace*{-4mm}
\label{table:change_from_sota_posetrack}
\end{table}

\begin{table}
\centering
\setlength\tabcolsep{1.0pt}
\def\arraystretch{1.1}
\begin{tabular}{L{1.3cm}C{0.8cm}C{0.8cm}C{0.8cm}C{0.8cm}C{0.8cm}C{0.8cm}C{0.8cm}C{0.8cm}}
\specialrule{.1em}{.05em}{.05em}
      Methods & Head & Shou & Elb & Wri & Hip & Knee & Ankl & Total \\ \hline
PoseRefiner~\cite{fieraru2018learning}    & 74.0 (-0.4) & 76.8 (-0.1) & 72.2 (+0.0) & 65.4 (+0.2) & 70.5 (+1.3) & 69.7 (-0.3) & 63.7 (+0.8) & 70.6 (+0.2) \\ 
\textbf{PoseFix (Ours)}      & \textbf{79.0 (+4.6)} & \textbf{81.6 (+4.7)} & \textbf{76.4 (+4.2)} & \textbf{69.7 (+4.5)} & \textbf{75.2 (+6.0)} & \textbf{74.3 (+4.3)} & \textbf{67.0 (+4.1)} & \textbf{75.0 (+4.6)}\\ \specialrule{.1em}{.05em}{.05em}
\end{tabular}
\vspace*{-3mm}
\caption{AP comparison between PoseRefiner~\cite{fieraru2018learning} and PoseFix on the PoseTrack 2018 validation set. The number in the parenthesis denotes the AP change from the input pose (\textit{i.e.}, Simple).}
\vspace*{-4mm}
\label{table:comparison_poserefiner}
\end{table}

\section{Qualitative results}
\begin{figure*}
\begin{center}
\includegraphics[width=0.95\linewidth]{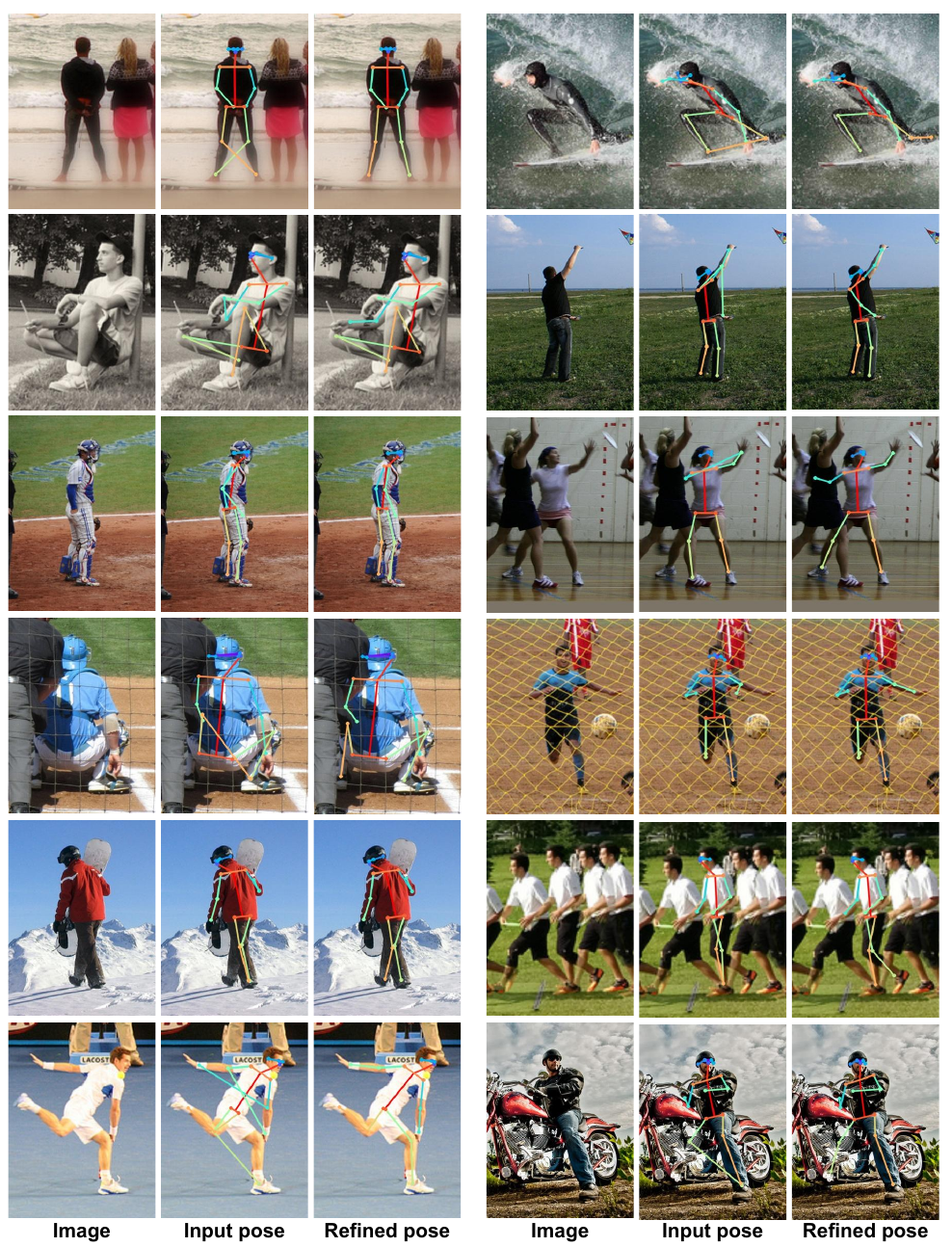}
\end{center}
\vspace*{-3mm}
   \caption{Qualitative results of the PoseFix on the test-dev set.}
\vspace*{-3mm}
\label{fig:q1}
\end{figure*}

\begin{figure*}
\begin{center}
\includegraphics[width=0.95\linewidth]{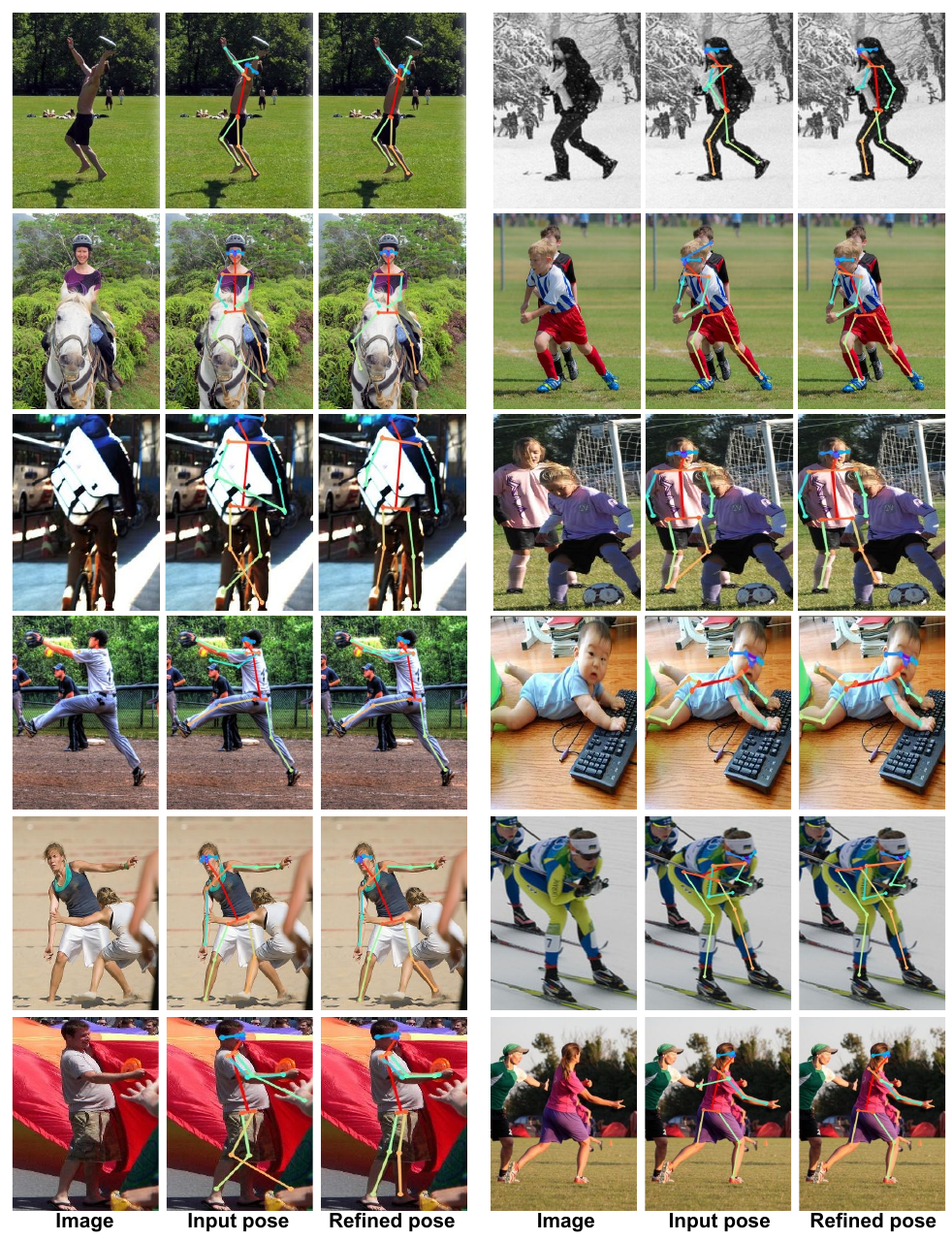}
\end{center}
\vspace*{-3mm}
   \caption{Qualitative results of the PoseFix on the test-dev set.}
\vspace*{-3mm}
\label{fig:q2}
\end{figure*}

\begin{figure*}
\begin{center}
\includegraphics[width=0.77\linewidth]{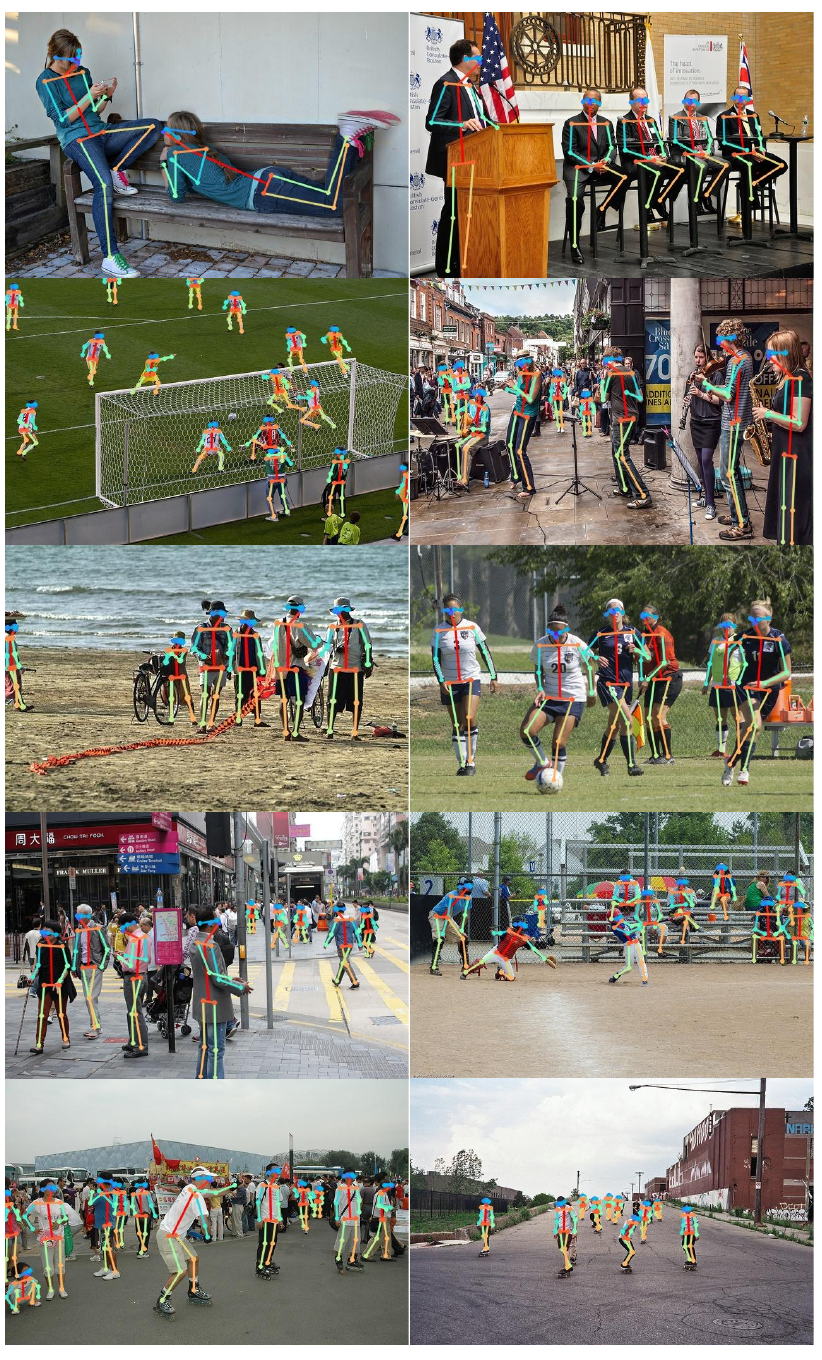}
\end{center}
\vspace*{-3mm}
   \caption{Qualitative results of the PoseFix on the test-dev set.}
\vspace*{-3mm}
\label{fig:q3}
\end{figure*}

We show some qualitative results on the MS COCO~\cite{lin2014microsoft} test-dev set. Figure~\ref{fig:q1} and ~\ref{fig:q2} show the input images, input poses, and refined poses when the PoseFix is applied to Mask R-CNN~\cite{he2017mask}. Figure~\ref{fig:q3} shows final results when the PoseFix is applied to Simple~\cite{xiao2018simple}.

\clearpage

{\small
\bibliographystyle{ieee}
\bibliography{bib/egbib}
}

\end{document}